\documentclass{article}
\usepackage{jfrExamplee}
\usepackage{graphicx}
\usepackage{apalike}
\usepackage{setspace}

\usepackage{amsmath,amssymb,mathrsfs}
\usepackage{upgreek}
\usepackage[ruled, linesnumbered, vlined]{algorithm2e}
\usepackage{comment}
\usepackage{graphicx}
\usepackage{graphics}
\usepackage[tight]{subfigure}
\usepackage{float}
\usepackage[table]{xcolor}
\usepackage{multirow}
\usepackage{array}
\usepackage{enumerate}
\usepackage{sidecap}
\usepackage{hyperref}
\usepackage{url}
\usepackage{wrapfig}
\usepackage{bm}
\usepackage{algpseudocode}
\graphicspath{{./figs/}{./}}
\usepackage{flushend}

\usepackage{apalike}

\usepackage{color}

\definecolor{lightblue}{rgb}{0,0.2,1}
\definecolor{black}{rgb}{0,0,0}
\hypersetup{
    colorlinks=true,%
    citebordercolor=white,%
    filebordercolor=white,%
    linkbordercolor=white,%
    urlbordercolor=white,%
    citecolor=black,%
    linkcolor=black,%
    urlcolor=black%
}

\newcounter{tecounter}
\DeclareMathOperator*{\argmin}{arg\,min}
\DeclareMathOperator*{\argmax}{arg\,max}

\title{Data-Driven Learning and Planning \\ for Environmental Sampling}

\author{
Kai-Chieh Ma \\
Department of Computer Science\\
University of Southern California\\
Los Angeles, CA 90089 \\
\texttt{kaichiem@usc.edu} \\
\And
Lantao Liu \\
Department of Intelligent Systems Engineering  \\
Indiana University\\
Bloomington, IN 47408 \\
\texttt{lantao@iu.edu} \\
\AND
Hordur K. Heidarsson \\
Department of Computer Science\\
University of Southern California\\
Los Angeles, CA 90089 \\
\texttt{heidarss@usc.edu} \\
\And
Gaurav S. Sukhatme \\
Department of Computer Science\\
University of Southern California\\
Los Angeles, CA 90089 \\
\texttt{gaurav@usc.edu} \\
}

\begin{document}

\maketitle


\begin{abstract}
Robots such as autonomous underwater vehicles (AUVs) and autonomous surface vehicles (ASVs) have been used for sensing and monitoring aquatic environments such as oceans and lakes.
Environmental sampling is a challenging task because the environmental attributes to be observed can vary
both spatially and temporally, and the target environment is usually a large and continuous domain whereas the sampling data is typically sparse and limited.  
The challenges require that the sampling method must be informative and efficient enough to catch up with the environmental dynamics.
In this paper we present a planning and learning method that enables a sampling robot to perform persistent monitoring tasks by learning and refining a  dynamic ``data map" that models a spatiotemporal environment attribute such as ocean salinity content. 
Our environmental sampling framework consists of two components:
to maximize the information collected, we propose an informative planning component that efficiently generates sampling waypoints that contain the maximal 
information;
To alleviate the computational bottleneck caused by large-scale data accumulated, we develop a component 
based on a sparse Gaussian Process whose hyperparameters are learned online by taking advantage of only a subset of data that provides the greatest contribution. 
We validate our method with both simulations running on real ocean data and field trials with an ASV in a lake environment.
Our experiments show that the proposed framework is 
both accurate in learning the environmental data map and efficient in  catching up with the dynamic environmental changes$^{\dagger}$.
\end{abstract}

\footnotetext{This work is an extension of a proceeding paper~\cite{ma2017icra}.}


\section{Introduction}

Scientists are able to gain a greater understanding of the environmental states (e.g., physical, chemical or biological parameters) through environmental sensing and monitoring~\cite{Dunbabin12}.
Typically, the environmental sensing involves a process of collecting data of important environmental attributes (e.g., temperature, salinity, pollution contents) at certain selected locations, and the goal is to build a ``data map" that can best describe the state of the environment. 

\begin{figure}[t]
 \centering
  \subfigure[Aug 15, 2016]
        {\label{fig:815}\includegraphics[height=1.5in]{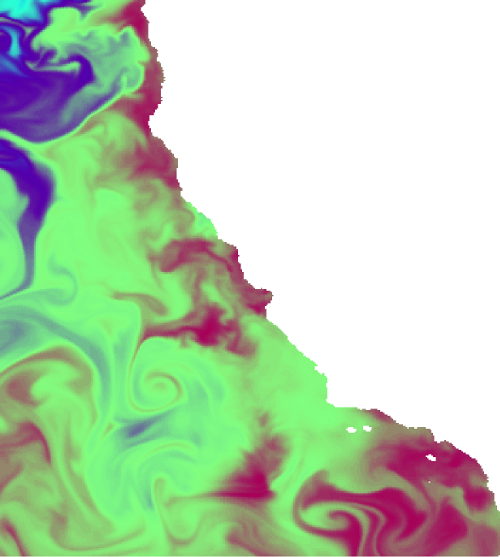}}
 \quad \quad
 \subfigure[Aug 22, 2016]
        {\label{fig:822}\includegraphics[height=1.5in]{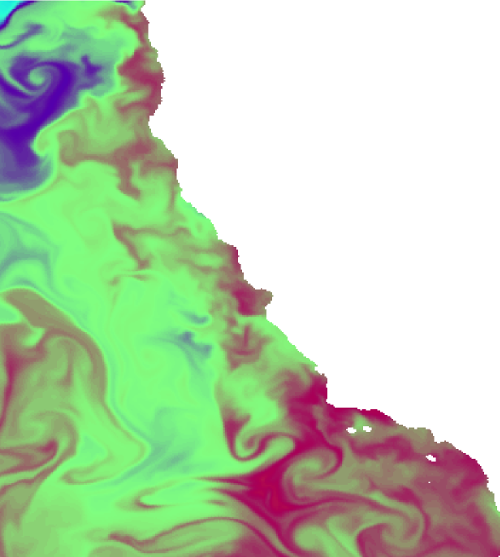}}
 \quad \quad 
 \subfigure[Aug 29, 2016]
        {\label{fig:829}\includegraphics[height=1.5in]{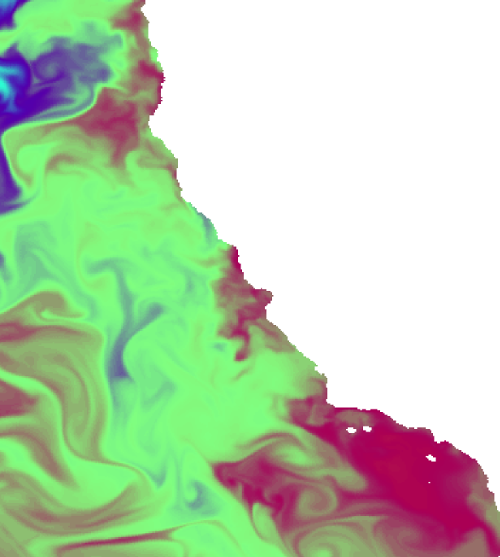}}
 \caption{Ocean salinity data in the Southern California Bight region generated by the Regional Ocean Modeling System (ROMS)~\cite{shchepetkin_regional_2005}. Color indicates levels of salinity content. 
 }
\label{fig:salinity} 
\end{figure}

However, the environment to be monitored is usually a large and continuous area whereas the sampled data is discrete and limited due to cost. 
In addition, 
the outdoor water environment is typically dynamic, so that any environmental attribute associated to water also varies as time elapses. 
Fig.~\ref{fig:salinity} illustrates the time-varying salinity data in the Southern California Bight region generated by the Regional Ocean Modeling System~(ROMS)~\cite{shchepetkin_regional_2005}.
The variations in spatial and temporal dimensions entail that 
the collected data must contain the maximal information in order to provide a good estimate of the environment at any time~\cite{Meliou07,Ouyang2014MAS}. 

Increasingly, a variety of autonomous robotic systems, including marine vehicles~\cite{Fiorelli03adaptivesampling}, aerial vehicles~\cite{classification12}, and ground vehicles~\cite{Trincavelli08}, are designed and deployed for environmental monitoring in order to replace  conventional methods 
of deploying static sensors to areas of interest~\cite{Oliveira11}.
Particularly, autonomous underwater vehicles (AUVs) such as marine gliders are becoming popular due to their long-range (hundreds of kilometers) and long-duration (weeks even months) monitoring capabilities~\cite{Miles2015,PaleyLeoZhang2006,LPDDLZ_JFR10}.

We are interested in estimating the current state of the environment and providing a nowcast (instead of forecast) of the environment, 
by navigating
robots to collect the information used for such estimation. 
To model spatial phenomena, a common approach is to use a rich class of techniques called Gaussian Processes (GPs) ~\cite{Rasmussen2005,Singh2007,Ouyang2014MAS}.
In this work, we also employ this broadly-adopted approach to build and learn an underlying dynamic ``data map" of a target attribute of interest.
More specifically, the data map is a scalar field/map where each point on the map represents either a sampled/known or a predicted value measuring a single type of environmental attribute of interest.
(Note, in general the environmental model may be complex and include multiple measuring and predicting attributes (variables), 
in our paper however, we use the term {\em environment model} and {\em data map} alternatively but they refer to the same predicted scalar field of single environmental attribute.)

Still, there are a few challenges to address: 
\begin{itemize}
\item 
The first challenge lies in the acquirement of the most useful sensing inputs for learning the model,
i.e., since samples from different locations are not equally important, thus we wish to seek for the samples that best describe the environment.
Navigating robots to obtain such samples is called {\em informative planning}~\cite{Singh2007,binney13}. 
In this work, we utilize the mutual information between visited locations and the remainder of the space to characterize the amount of information (information gain) collected. 

\item
The second challenge is the relaxation of the prohibitive computational cost for maintaining the model. 
The most accurate way to estimate a latent model is to use all historical sensing data. 
However, since the environmental monitoring task can be long-range and long-term, the data size continuously grows and it will eventually ``explode". 
Consequently, an efficient estimator will need to dynamically select only the most informative data while abandoning the samples that contribute less.
    
\end{itemize}

We propose an informative planning and online learning approach for long-term environmental monitoring. 
Our objective is to construct and maintain an environmental model by navigating a robot to collect data with the greatest information. 

This paper includes the following contributions:
(1) We propose an environmental monitoring framework that allows online estimate of certain spatiotemporal environmental state and the associated data map. Different from most existing works where only spatial models need to be constructed, in our work we also consider the dynamic process of environment. In order to obtain an accurate estimate at any time, the dynamic environmental changes require online data processing and model learning.
We propose a framework that discretizes the time into ``piecewise static" intervals to approximate the temporal dynamics. By connecting and extending a sparse variant of Gaussian Process to such temporal model, 
we show that the real-time environment estimate becomes possible.
(2) We consider the efficiency of constructing the data map and aim at building an accurate map as fast as possible.
Based on the prediction confidence produced from the Gaussian Process, we develop an information-theoretic planning framework where the robot is guided to explore those regions that are most uncertain.
We utilize the mutual information between the known space and unknown space to quantify the information and have 
transformed the computation process into an efficient dynamic programming structure. This enables us to compute the future sampling locations that are most informative, so that the unknown or uncertain space can be explored quickly.
(3) We have validated the proposed framework with both simulations and field trials, and compared our method with other popular baseline approaches such as the lawnmower and Monte Carlo random sampling strategies. The results revealed that our approach is the fastest in estimating the spatiotemporal data map to reach given accuracy.  



\section{Related Work}

Environmental sensing and monitoring allow scientists to assess the processes of a particular environment, and have been used in a broad range of applications. For example, an array of sensor networks can be deployed to detect forest fires~\cite{lloret2009wireless} and volcano activities~\cite{werner2006fidelity};
unmanned aerial vehicles (UAVs) have been used to estimate yields of crops or fruits~\cite{nuske2011yield,yang2012high} and to study spatial ecology and its spatiotemporal dynamics~\cite{anderson2013lightweight};
with a capacity of performing long-range and long-term tasks, marine robots can collect large-area ocean data~\cite{smith2011persistent} and trace chemical plumes~\cite{farrell2005chemical,hajieghrary2015information}; autonomous boats have
been used to monitor fish schools~\cite{tokekar2010robotic}.

Methodologies for environmental sensing and monitoring have also been developed to address challenging problems in different scenarios.
For instance, stochastic search has been considered for switching fields~\cite{AzSaSePa2012}, 
and Markov Chain Monte Carlo (MCMC) techniques have been used to study convergence~\cite{HuHoChLy2012}.
Convergence guarantees were also analyzed for control using gradient-aware straight line motions for point-mass vehicles~\cite{MaSaTe2007}.
Distributed control methods based on gradient climbing~\cite{OrFiLe2004}, consensus~\cite{Co2007}, and
source seeking under limited communication~\cite{LiGu2012}, have been developed using different optimization techniques. 
Further work using multiple vehicles~\cite{MaBoPe2010} took account of simplified turbulent flows~\cite{ChWeWeWeZh2013} and
bio-inspired methods with stability guarantees~\cite{WuChZh2013}.

In artificial intelligence and robotics research domain, the planning and environment monitoring are two big and well studied topics. 
Here we focus on reviewing the works that are related to the informative planning and the GP-based environmental model prediction as well as the sparse variants of GPs.
Formally, the planning method that navigates robots to collect and maximize information gain through sampling the environment is called {\em informative planning}~\cite{Singh2007,binney13}. 
A variety of methodologies have been proposed to tackle the informative planning problem, among which the most investigated approaches belong to the {\em nonmyopic} framework.   
Formally, the term {\em myopic} means that the path waypoints are computed independently and greedily, without considering the cost and consequences of making observations in a long horizon of future. Instead, a nonmyopic strategy performs optimization and computes a series of waypoints by considering the effect of later time-steps~\cite{Meliou07}.

Representative nonmyopic informative planning approaches include, for example, a {\em recursive-greedy} based algorithm~\cite{Singh2007} where the informativeness is generalized as a {\em submodular set function}  built on which a sequential-allocation mechanism is designed in order to obtain subsequent waypoints. 
This recursive-greedy framework has been extended by taking into account the avoidance of shipping lanes~\cite{Binney-2010-642} and diminishing returns~\cite{binney13}. 
Differing from above mechanisms where the path waypoints are built by separate searching techniques with the informativeness as some utility function, Low~\cite{Low2009thesis} proposed a differential entropy based planning method in which a batch of waypoints can be obtained through solving a dynamic program.
Such a framework has been extended to approaches incorporating mutual information~\cite{Cao2013} and Markovian~\cite{Low2011} optimization criteria.
However, these approaches are formulated with an assumption that the underlying map is in a regular shape (e.g., a parallelogram shape) 
and the map is transected (sliced) column-wise, so that each algorithmic iteration computes waypoints within a column and the navigation paths are obtained by connecting those waypoints among the pairwise adjacent columns.  
We recently proposed a informative planning method based on the dynamic programming structure in order to compute the informative waypoints across an arbitrary continuous space~\cite{ma2016information}. This nonmyopic method has also been combined with Markov Decision Process to cope with robot's action uncertainty caused by external disturbances. 
In addition, there are also many methods optimizing over complex planning and information constraints (e.g., \cite{SolteroSR12,YuSchRus14ICRA}).
Also note that, although this paper does not discuss much on the vehicle's low level motion, the motion control is very important in practice, especially if there are external disturbances.
The vehicle's motion under ocean current disturbance can be solved with different theoretical frameworks, for instance, 
it can be formulated under the control framework (e.g., see~\cite{mukhopadhyay2014collaborative}) or the decision-theoretic planning framework (e.g., see~\cite{fackler2014monitoring}).

To model spatial phenomena in a continuous domain, a broadly adopted method is the Gaussian Process which is a generic supervised learning method designed to solve regression and probabilistic classification problems~\cite{Rasmussen2005,Singh2007,Ouyang2014MAS}.
The regression property of GP has manifested itself as 
a powerful tool for predicting environment states based on which the environment can be monitored.
In the geostatistics or spatial statistics literature, the GP regression technique is usually called {\em kriging}~\cite{hengl2009practical} and is mostly used to analyze spatial properties (
oftentimes kriging relies on the knowledge of certain spatial
structure, which is modeled via the second-order properties, i.e. variogram or covariance,
of the underlying random function~\cite{lichtenstern2013kriging}).
In this work, although GP is employed to predict a data map representing the spatio-temporal environment, we rather treat it as a kernel based machine learning method where its Bayesian learning property is exploited with online collected data. 

There are many applications where GPs have been utilized as a basic framework to model various environments.
For instance, GP has been used to design placement patterns of static sensors in a sensor network so that the environment model can be predicted with a solution that is near-optimal~\cite{guestrin2005near}. 
Optimization of static sensor placement has been applied for modeling indoor 3-dimensional environment~\cite{multi-sensor2015} and outdoor urban environment~\cite{gps2016urban} through appropriate kernels (covariance functions).
In addition to the applications in sensor networks, GPs are also being exploited in robotics research.
For example, GP has been used on a mobile robot to build a spatial model describing gas distribution~\cite{stachniss2008gas}, or provide a measure of uncertainty to guide sensor-centric robot localization~\cite{brooks2008gaussian};
through combining with mutual information and Bayesian optimization, GP 
has been utilized to guide robots to explore unknown  static environment~\cite{bai2016}. 
In dynamic environment settings, variants of GPs have also been employed to learn uncertainty
models of ocean processes to assist in the operation of AUVs in the ocean~\cite{hollinger2013learning,hollinger2016learning}. 
By integrating with a deterministic vehicle routing scheme, we also proposed a method for informative ocean sampling and monitoring in dynamic ocean environments with multi-robot systems~\cite{ma2016DARS}.

This presented work is also built on top of a GP which is used to describe an interested environmental attribute (e.g., we measure the salinity and build a continuous salinity data-map).
Our GP model here 
is coupled with the information-theoretic framework where we compute the mutual information
between visited locations and the remainder of the space
to characterize the amount of information collected, through which we use the informative planning framework to compute high-level informative navigation waypoints. 
Different from methods mentioned above, in this work we consider dynamic environments that vary both spatially and temporally. 
This means that the latent parameters that support/characterize the GP model need to be updated too in order to accurately reflect the ground-truth environment in a timely manner. 

To improve GP's 
prediction accuracy, the choice over different prior covariance functions and the update of its hyperparameters are crucial, especially in the scenarios involving many spatiotemporal dynamics. 
This problem is typically referred as {\em Model Selection and Adaptation of Hyperparameters}~\cite{Rasmussen2005}. Particularly, the adaptation of hyperparameters can be updated using a data-driven approach. The most common ones are done by maximizing the marginal likelihood or minimizing the generalization error using cross-validation approach. For the case of GP classification, other optimization criteria such as {\em Alignment}~\cite{ello2001kernel} can also be adopted.

A critical problem for persistent (long-term even life-long) tasks that one must consider is the large-scale data accumulated. Although abundant 
data might predict the most accurate model, in practice large amounts of data can exceed the robot's onboard computational capacity. 
Methods for reducing the computing burdens of GPs have been previously investigated. For example, GP regressions can be done in a real-time fashion where the problem can be estimated locally with local data~\cite{Nguyen-tuong08localgaussian}.
Another representative framework is a sparse representations of the GP model~\cite{csato2002sparse,smola2001sparse} which is based on a combination of a Bayesian online algorithm together with a sequential construction of the most relevant subset of the data. 
This method allows the model to be refined in a recursive way as the data streams in. 
The framework has been further extended to many application domains such as visual tracking~\cite{Ranganathan11} and spatial modeling~\cite{stachniss2008gas}. 

Recently, we proposed an informative planning and online learning approach for long-term spatiotemporal environmental monitoring~\cite{ma2017icra} with the objective to construct and maintain a spatiotemporal environmental model by navigating the robot to the most informative regions to collect data with the greatest information. 
The method has integrated a sparse variant of GPs so that both the model and hyperparameters can be improved online with dynamic but a fixed size of data. 
Then the ameliorated environment model is in turn used to improve the planning component at appropriate re-planning moments. 
This paper is an extension version of the the previous work~\cite{ma2017icra} and distinguishes itself from earlier work in the following aspects.
First, we have extended the work by adding many necessary details so that the method is theoretically complete and in-depth.
Second, we have added new simulation results to capture data (environmental attribute) variations in the temporal dimension.
Third, we have validated the approach in field trials, where we implemented our method on an ASV developed in our lab, and have deployed it in a lake to collect data and estimate a continuous data 
map of the lake.
Both simulation and field trial experiments show that our method is accurate and fast in learning a spatiotemporal environmental model.


\section{Preliminaries}

In this section, we briefly present the preliminary background for the GP-based environmental modeling.

\subsection{Gaussian Process Regression on Spatial Data} \label{sec:gp}

A Gaussian Process is defined as a collection of random variables where any finite number of which have a joint Gaussian distribution. 
GP's prediction behavior is determined by the prior covariance function (also known as {\em kernel}) and the training points.
The prior covariance function describes the relation between two independent data points and it typically comes with some free hyperparameters to control the relation.

Formally, let $X$ be the set of $n$ training points associated with target values, $\bm{y}$, and let $X_*$ be the testing points. The predictive equations
of the GP regression can be summarized as:
\begin{equation} \label{eq:gp}
    \begin{aligned}
        \bm{f}_* | X, \bm{y}, X_* &\sim \mathcal{N}(\bar{\bm{f}_*}, \textrm{cov}(\bm{f}_*)) \\
        \bar{\bm{f}_*} &\triangleq \mathbb{E}[\bm{f}_* | X, \bm{y}, X_*] = K(X_*, X)K(X, X)^{-1} \bm{y} \\
		\textrm{cov}(\bm{f}_*) &= K(X_*, X_*) - K(X_*, X)K(X, X)^{-1}K(X, X_*)
	\end{aligned}
\end{equation}
where $K(\cdot, \cdot)$ denotes a matrix where each entry is the covariance evaluated by the kernel function. For example, $K(X, X_*)$ is a $|X| \times |X_*|$ matrix evaluated by a pre-selected kernel function for all pairwise data points in $X$ and $X_*$.
A widely adopted choice of kernel function for spatial data is the {\em squared exponential automatic relevance determination} function~\cite{neal1996bayesian}:
\begin{equation}
	\begin{split}
	k(\bm{x}, \bm{x}') = \sigma_f^2 \exp(-\frac{1}{2}(\bm{x}-\bm{x}')^TM(\bm{x}-\bm{x}')) + \sigma_n^2\delta_{xx'} 
	\end{split}
\end{equation}
where
$M = diag(\bm{l})^{-2}$. The parameters $\bm{l}$ are the {\em length-scales} in each dimension of $\bm{x}$ and determine the level of correlation (each $l_i$ models the degree of smoothness in the spatial variation of the measurements in the $i$th dimension of the feature vector $\bm{x}$). $\sigma_f^2$ and $\sigma_n^2$ denote the variances of the signal and noise, respectively. $\delta_{xx'}$ is the Kronecker delta function which is 1 if $\bm{x}=\bm{x}'$ and zero otherwise.

\subsection{Estimation of Hyperparameters Using Training Data}\label{sec:HyperTraining}
Let $\bm{\theta} \triangleq \{\sigma_n^2, \sigma_f^2, \bm{l}\}$ be the set of hyperparameters in the kernel function. We are interested in estimating these hyperparameters so that the kernel function can describe the underlying phenomena as accurate as possible.
A common approach to learning the set of hyperparameters is via maximum likelihood estimation combined with $k$-fold cross-validation (CV)~\cite{Rasmussen2005}. An extreme case of the $k$-fold cross-validation is when $k$ is equal to the number of training points $n$. 
It is also known as the {\em leave-one-out cross-validation} (LOO-CV). 
Mathematically, the log-likelihood for the leaving out training case $i$ is
\begin{equation} \label{eq:log-likelihood}
	\log p(y_i | X, \bm{y}_{-i}, \bm{\theta}) = -\frac{1}{2}\log [\textrm{var}(\bar{f_i})] - \frac{(y_i - \bar{f_i})^2}{2\textrm{var}(\bar{f_i})} - \frac{1}{2}\log(2\pi)
\end{equation}
where $\bm{y}_{-i}$ denotes all targets in the training set except the one with index $i$, and $\bar{f_i}$ and $\textrm{var}(\bar{f_i})$ are calculated according to Eq.~\eqref{eq:gp}. The log-likelihood of LOO is therefore
\begin{equation}
	L_{LOO}(X, \bm{y}, \bm{\theta}) = \sum_{i = 1}^{n}\log p(y_i | X, \bm{y}_{-i}, \bm{\theta}).
\end{equation}
Notice that in each of the $|\bm{y}|$ LOO-CV iterations, the inverse of the training set covariance matrix, $K(X_{-i}, X_{-i})^{-1}$, is needed, which is costly if computed repeatedly. This can actually be computed efficiently from the inverse of the complete covariance matrix using {\em inversion by partitioning}~\cite{press1996numerical}. The resulting predictive mean and variance can then be formulated as
\begin{equation} \label{eq:cv}
    \begin{aligned}
        \bar{f_i} &= y_i - [K(X, X)^{-1}\bm{y}]_i / [K(X, X)^{-1}]_{ii} \\
        \textrm{var}(\bar{f_i}) &= 1 / [K(X, X)^{-1}]_{ii}
	\end{aligned}
\end{equation}

To obtain the optimal values of hyperparameters $\bm{\theta}$, we can compute the partial derivatives of $L_{LOO}$ and use the gradient optimization techniques.
The partial derivatives of $L_{LOO}$ using chain rules is
\begin{equation} \label{eq:L_LOO}
    \begin{aligned}
        \frac{\partial L_{LOO}}{\partial \theta_j} &= \sum_{i = 1}^{n} \Big( \frac{\partial \log p(y_i | X, \bm{y}_{-i}, \bm{\theta})}{\partial \bar{f_i}}\frac{\partial \bar{f_i}}{\partial \theta_j} + \frac{\partial \log p(y_i | X, \bm{y}_{-i}, \bm{\theta})}{\partial \textrm{var}(\bar{f_i})}\frac{\partial \textrm{var}(\bar{f_i})}{\partial \theta_j}
        \Big)\\
	\end{aligned}
\end{equation}
To calculate the partial derivatives of $L_{LOO}$, we need the partial derivatives of the LOO-CV predictive mean and variances from Eq.~\eqref{eq:cv}.
\begin{equation} 
    \begin{aligned}
        \frac{\partial \bar{f_i}}{\partial \theta_j} &= \frac{[Z_j\bm{\alpha}]_i}{[K(X, X)^{-1}]_{ii}} - \frac{\bm{\alpha}_i[Z_j K(X, X)^{-1}]_{ii}}{[K(X, X)^{-1}]_{ii}^2} \\
        \frac{\partial \textrm{var}(\bar{f_i})}{\partial \theta_j} &= \frac{[Z_j K(X, X)^{-1}]_{ii}}{[K(X, X)^{-1}]_{ii}^2}
	\end{aligned}
\end{equation}
where $\bm{\alpha} = K(X, X)^{-1}\bm{y}$ and $Z_j = K(X, X)^{-1}\frac{\partial K(X, X)}{\partial \theta_j}$. \\
Substitute $\frac{\partial \bar{f_i}}{\partial \theta_j}$ and $\frac{\partial \textrm{var}(\bar{f_i})}{\partial \theta_j}$ in Eq.~\eqref{eq:L_LOO} and calculate the partial derivative from Eq.~\eqref{eq:log-likelihood}, we have
\begin{equation}
    \begin{aligned}
        \frac{\partial L_{LOO}}{\partial \theta_j} = \sum_{i = 1}^{n} &\frac{1}{[K(X, X)^{-1}]_{ii}} \Big(\alpha_i[Z_j\bm{\alpha}]_i  -\frac{1}{2}(1 + \frac{\alpha_i^2}{[K(X, X)^{-1}]_{ii}}) [Z_j K(X, X)^{-1}]_{ii}\Big),
	\end{aligned}
\end{equation}
Using the standard gradient method, we update each $\theta_j$ iteratively:
\begin{equation} 
    \theta_j^{(t+1)} = \theta_j^{(t)} + \eta \frac{\partial L_{LOO}}{\partial \theta_j^{(t)}},
\end{equation}
where $\eta$ is the learning rate.


\section{Online Learning and Informative Planning}

As aforementioned, one limitation of GPs for a long-term mission is the memory requirement for large (possibly infinite) training sets. 
In our system, we borrow the idea of Sparse Online Gaussian Process (SOGP)~\cite{csato2002sparse} to overcome this limitation. The method is based on a combination of a Bayesian online algorithm together with a sequential subsampling of the data which best describes a latent model.

\subsection{Bayesian Learning with Gaussian Processes} \label{sec:OnlineGP}
Given a prior GP $\hat{p}_t(\bm{f})$ at time $t$, when a new data point $(\bm{x}_{t+1}, y_{t+1})$ at time $t+1$ comes in, it is  incorporated by performing a Bayesian update to yield a posterior.
\begin{equation}
	p_{post}(\bm{f}) = \frac{p(y_{t+1}|\bm{f}) \hat{p}_t(\bm{f})}{\mathbb{E}_{\hat{p}_t(\bm{f})}[p(y_{t+1}|\bm{f}_{D})]},
\end{equation}
where $\bm{f} = [f(\bm{x}_1), \dots, f(\bm{x}_M)]^{T}$ denotes a set of function values, and $\bm{f}_{D} \subseteq \bm{f}$ where $\bm{f}_{D}$ is the set of $f(\bm{x}_i) = f_i$ with $\bm{x}_i$ in the training set. In general, $p_{post}(\bm{f})$ is no longer Gaussian unless the likelihood itself is also Gaussian. 
Therefore, $p_{post}(\bm{f})$ is projected onto the closest GP, $\hat{p}_{t+1}$ where $\hat{p}_{t+1} = \argmin_{\hat{p}}$ KL$(p_{post}(\bm{f}) || \hat{p})$.
(KL is the Kullback-Leibler divergence that is used to measure the difference between two probability distributions.) 
It is shown in ~\cite{Opper:1999:BAO:304710.304756} that the projection results in a good matching of the first two moments (mean and covariance) of $p_{post}$ and the new Gaussian posterior $\hat{p}_{t+1}$. By following the lemma of~\cite{csato2002sparse}, we arrive at the parametrization for the approximate posterior GP at time $(t+1)$ as a function of the kernel and likelihoods. 
\begin{equation} \label{eq:SOGP_Posterior}
    \begin{aligned}
        \bar{f_*} &= \sum_{i=1}^{t+1} k(\bm{x}_*, \bm{x}_i) \alpha_{t+1}(i) = \bm{\alpha_{t+1}}^T\bm{k}_{\bm{x}_*, {t+1}} \\
        \textrm{var}(f_*) &= k(\bm{x}_*, \bm{x}_*) + \sum_{i, j = 1}^{t+1} k(\bm{x}_*, \bm{x}_i)[C_{t+1}]_{ij}k(\bm{x}_j, \bm{x}_*) \\
                &\triangleq k(\bm{x}_*, \bm{x}_*) + \bm{k}_{\bm{x}_*, t+1}^T C_{t+1} \bm{k}_{\bm{x}_*, t+1}
	\end{aligned}
\end{equation}
where $\bm{k}_{\bm{x}_*, t+1} = [k(\bm{x}_1, \bm{x}_*), \dots, k(\bm{x}_{t+1}, \bm{x}_*)]^T$, and $\bm{\alpha_{t+1}}$ and $C_{t+1}$ are updated using
\begin{equation}
    \begin{aligned}
        \bm{\alpha_{t+1}} &= T_{t+1}(\bm{\alpha_t}) + q_{t+1} \bm{s}_{t+1} \\
        C_{t+1} &= U_{t+1}(C_t) + r_{t+1} \bm{s}_{t+1} \bm{s}_{t+1}^T \\
        \bm{s}_{t+1} &= T_{t+1}(C_t \bm{k}_{\bm{x}_*, t+1}) + \bm{e}_{t+1} \\
        q_{t+1} &= \frac{\partial}{\partial\mathbb{E}_{\hat{p}_t(\bm{f})}[\bm{f}_{t+1}]} \log\mathbb{E}_{\hat{p}_t(\bm{f})}[p(y_{t+1} | \bm{f}_{t+1})] \\
        r_{t+1} &= \frac{\partial^2}{\partial\mathbb{E}_{\hat{p}_t(\bm{f})}[\bm{f}_{t+1}]^2} \log\mathbb{E}_{\hat{p}_t(\bm{f})}[p(y_{t+1} | \bm{f}_{t+1})] \\
	\end{aligned}
\end{equation}
where $\bm{e}_{t+1}$ is the $(t+1)$-th unit vector. The operator $T_{t+1}$ ($U_{t+1}$) is defined to extend a $t$-dimensional vector (matrix) to a $(t+1)$-dimensional one by appending zero at the end of the vector (zeros at the last row and column of the matrix).
Initially at time $0$, $\bm{\alpha_0}$ and $C_0$ are set to be zero-sized vector and matrix.
For the regression with Gaussian noise (variance $\sigma_0^2$), The expected likelihood is a normal distribution with mean $\bar{f_*}$ and variance $\textrm{var}(f_*) + \sigma_0^2$. Hence, the logarithm of the expected likelihood is:
\begin{equation}
  \begin{split}
    \log\mathbb{E}_{\hat{p}_{t-1}(\bm{f})}[p(y_t | \bm{f}_t)] &= -\frac{1}{2}\log[2\pi(\textrm{var}(f_*) + \sigma_0^2)] 
    - \frac{(y_t - \bar{f_*})^2}{2(\textrm{var}(f_*) + \sigma_0^2)},
  \end{split}
\end{equation}
and the first and second derivatives with respect to the mean $\bar{f_*}$ are
\begin{equation}
    \begin{aligned}
        q_t &= \frac{y_t - \bar{f_*}}{\textrm{var}(f_*) + \sigma_0^2}, \\
        r_t &= -\frac{1}{\textrm{var}(f_*) + \sigma_0^2}, \\
	\end{aligned}
\end{equation}
where both $q_t$ and $r_t$ are scalars.


\subsection{Online Process of Sparse Samples} \label{sec:SparseGP}

To prevent the unbounded growth of memory requirement due to the increase of data, it is necessary to limit the number of the training points which are stored in a {\em basis vector set (BV-set)}, while preserving the predictive accuracy of the model. This is done in two different stages. 

First, when a new training point $(\bm{x}_{t+1}, y_{t+1})$ at time $t+1$ arrives, if there exists a $\hat{\bm{e}}_{t}$ such that the relation
\begin{equation}
    k(\bm{x}, \bm{x}_{t+1}) = \sum_{i=1}^{t} \hat{\bm{e}}_{t+1}(i)k(\bm{x}, \bm{x}_{i})
\end{equation}
holds for all $\bm{x}$ in the input space, it essentially means the feature vector $\Upphi(\bm{x}_{t+1})$ lies exactly on the space spanned by the current BV-set, $\big(\Upphi(\bm{x}_{1}), \dots, \Upphi(\bm{x}_{t})\big)$, where $\Upphi$ is a transformation function that transforms the input $\bm{x}$ into a feature space. If this is the case, then the GP can still be modeled using only the first $t$ inputs, but with ``re-normalized" parameters $\hat{\bm{\alpha}}_{t+1}$ and $\hat{C}_{t+1}$, which can be done by updating $\hat{\bm{s}}_{t+1}$ via: 
\begin{equation}\label{eq:approximation_update}
    \hat{\bm{s}}_{t+1} = C_t \bm{k}_{\bm{x}_{t+1}, t} + \hat{\bm{e}}_{t+1}.
\end{equation}
However, the exact $\hat{\bm{e}}_{t+1}$ does not always exist for most kernels and inputs $\bm{x}_{t+1}$. Nevertheless, we could try to approximate it by minimizing the error measure
\begin{equation}\label{eq:error_measure}
    || k(\bm{x}, \bm{x}_{t+1}) - \sum_{i=1}^{t} \hat{\bm{e}}_{t+1}(i)k(\bm{x}, \bm{x}_{i}) ||^2,
\end{equation}
where $||\cdot||$ is the norm operation. The minimization (~\cite{csato2002sparse}) of Eq.~\eqref{eq:error_measure} leads to 
\begin{equation}
    \hat{\bm{e}}_{t+1} = Q_{t}^{-1} \bm{k}_{\bm{x}_{t+1}, t}, 
\end{equation} 
where $Q_t = K(X_{t}, X_{t})^{-1}$ is the inversion of the full kernel matrix.
Note that the re-normalization update in Eq.~\eqref{eq:approximation_update} will be done only if the approximation error does not exceed some predefined threshold $\omega$, otherwise, the sample is added into the BV-set as described in \ref{sec:OnlineGP}. Let the quantity of the approximation error be $\gamma_{t+1}$. Specifically, it is the squared norm of the ``residual vector" from the projection in the feature space spanned by the current BV-set.
\begin{equation}
    \gamma_{t+1} = k(\bm{x}_{t+1}, \bm{x}_{t+1}) - \bm{k}_{\bm{x}_{t+1}, t}^T Q_{t} \bm{k}_{\bm{x}_{t+1}, t},
\end{equation}
Essentially, $\gamma_{t+1}$ can also be thought of as a form of ``novelty" for the new training point $(\bm{x}_{t+1}, y_{t+1})$. 


Second, when the size of BV-set exceeds the memory limit (or any pre-defined limit), $m$, a score measure is used to pick out the lowest one and remove it from the existing BV-set. Formally, let $\epsilon_i$ be the scoring function for the $i$th element in the BV-set, 
\begin{equation}
    \epsilon_i = \frac{|[\bm{\alpha}_{t+1}]_i|}{[Q_{t+1}]_{ii}}
\end{equation}
which is a measure of change on the expected posterior mean of a sample due to sparse approximation.
Assume the $j$th element in BV-set is the one with the lowest $\epsilon$, the removal of it requires a re-update of parameters $\bm{\alpha}_{t+1}$, $C_{t+1}$ and $Q_{t+1}$
\begin{equation}
    \begin{aligned}
        \hat{\bm{\alpha}}_{t+1} &= \bm{\alpha}^{(t)} - \alpha^j \frac{Q^j}{q^j} \\
        \hat{C}_{t+1} &= C^{(t)} + c^j\frac{Q^jQ^{jT}}{q^{j2}} - \frac{1}{q^j}[Q^jC^{jT} + C^{j}Q^{jT}] \\
        \hat{Q}_{t+1} &= Q^{(t)} - \frac{Q^{j} Q^{jT}}{q^j},
	\end{aligned}
\end{equation}
where $C^{(t)}$ is the resized matrix by removing the $j$th column and the $j$th row from
$C_{t+1}$, $C^j$ is the $j$th column of $C_{t+1}$ excluding the $j$th element and $c^j = [C_{t+1}]_{jj}$. 
Similar operations apply for $Q^{(t)}$, $Q^{j}$, $q^{j}$, $\bm{\alpha}^{(t)}$,  and $\alpha^j$.


\subsection{Environment Representation and Informative Sampling} \label{sec:infoPlanner}


To facilitate the computation of future informative sampling locations, we discretize the environment into a grid map where each grid represents a possible sampling location. 
The mean and variance of the measurement at each grid can be predicted via the SOGP model. 
We use the {\em mutual information} between the visited locations and the remainder of the space to characterize the
amount of information (information gain) collected.
Formally, the mutual information between two sets of random variables of measurements, $Z_A$, $Z_B$ sampled at locations, $A$, $B$ respectively, can be evaluated as:
\begin{equation}
    I(Z_A;Z_B) = I(Z_B; Z_A)= H(Z_A) - H(Z_A|Z_B).
\end{equation}
The entropy $H(Z_A)$ and conditional entropy $H(Z_A|Z_B)$ can be calculated by 
\begin{equation}
    \begin{aligned}
        H(Z_A) &= -\int p(Z_A)\log p(Z_A)d(Z_A) 
            = \frac{1}{2}\log\Big((2 \pi e)^k |\Sigma_{AA}|\Big) \\
	    H(Z_A|Z_B) &= \frac{1}{2}\log\Big((2 \pi e)^k|\Sigma_{A|B}|\Big) \\
	\end{aligned}
\end{equation}
where $k$ is the size of $A$. The covariance matrix $\Sigma_{AA}$ and $\Sigma_{A|B}$ can essentially be calculated from the posterior GP described in Eq.~\eqref{eq:SOGP_Posterior}.

To compute the future sampling locations, let $X$ denote the entire sampling space (all grids), and $Z_{X}$ be measurements for data points in $X$.
The objective is to find a subset of sampling points, $P\subset X$ with a size $|P| = n$, which gives us the most information for predicting our model.
This is equivalent to the problem of finding new sampling points in the un-sampled space that maximizes the mutual information between sampled locations and un-sampled part of the map.
The optimal subset of sampling points, $P^*$, with maximal mutual information is
\begin{equation}\label{eq:objective}
	P^* = \argmax_{P\in\mathcal{X}}I(Z_P;Z_{X \setminus P})
\end{equation}
where $\mathcal{X}$ represents all possible combinatorial sets, each of which is of size $n$. $P^*$ can be computed efficiently using a dynamic programming (DP) scheme~\cite{ma2016information}. 
In greater detail,
let $\bm{x}_i \in X$ denote an arbitrary sampling point at DP stage $i$ and $\bm{x}_{a:b}$ represent a sequence of sampling points from stage $a$ to stage $b$. 
The mutual information between the desired sampling points (which eventually form $P$) and the remaining map can then be written as $I(Z_{\bm{x}_{1:n}};Z_{X \setminus \{\bm{x}_{1:n}\}})$, which can be expanded using the chain rule:
\begin{equation}
	\begin{split}
		I(Z_{\bm{x}_{1:n}}; &Z_{X \setminus \{\bm{x}_{1:n}\}}) = I(Z_{\bm{x}_{1}}; Z_{X \setminus \{\bm{x}_{1:n}\}}) 
		+ \sum_{i=2}^{n}I(Z_{\bm{x}_{i}}; Z_{X \setminus \{\bm{x}_{1:n}\}}|Z_{\bm{x}_{1:i-1}}). 
	\end{split}
\end{equation}
One can utilize this form of mutual information to calculate $\bm{x}_i$ step by step, however, at every stage $i$ before the final stage, the entire unobserved set $X \setminus \{\bm{x}_{1:n}\}$ is not known in advance, therefore we make an approximation
\begin{equation} \label{eq:mutual}
	\begin{split}
		I(Z_{\bm{x}_{1:n}}; &Z_{X \setminus \{\bm{x}_{1:n}\}}) \approx I(Z_{\bm{x}_1}; Z_{X \setminus \{\bm{x}_1\}}) 
		 + \sum_{i=2}^{n}I(Z_{\bm{x}_i}; Z_{X \setminus \{\bm{x}_1, \dots, \bm{x}_i\}}|Z_{\bm{x}_{1:i-1}}),
	\end{split}
\end{equation}
Eq.~\eqref{eq:mutual} can now be expressed in a recursive form, i.e. for stages $i = 2, \dots, n$, the value $V_i(\bm{x}_i)$ of $\bm{x}_i$ is:
\begin{equation*}
    \begin{split}
        V_i(\bm{x}_i) = &\max_{\bm{x}_i \in X \setminus \{ \bm{x}_1, \dots, \bm{x}_{i-1}\}}I(Z_{\bm{x}_i}; Z_{X \setminus \{ \bm{x}_1, \dots, \bm{x}_i\}}|Z_{\bm{x}_{1:i-1}}) 
            + V_{i-1}(\bm{x}_{i-1}),
    \end{split}
\end{equation*}
with a recursion base case 
$
	V_1(\bm{x}_1) = I(Z_{\bm{x}_{1}}; Z_{X \setminus \{\bm{x}_1\}}).
$
Then with the optimal solution in the last stage, $\bm{x}_{n}^* = \argmax_{\bm{x}_{n} \in X}V_{n}(\bm{x}_{n})$, we can backtrace all optimal sampling points until the first stage $\bm{x}_1^*$, and obtain $P^*=\{\bm{x}^*_{1}, \bm{x}^*_{2}, \dots , \bm{x}^*_{n}\}$. The whole computational process for the information-driven planner is pseudo-coded in Alg.~\ref{algo:InfoPlanner}.

Note that, the informativeness maximization procedure only outputs batches of sampling points, but does not convey any information of ``a path" which is a sequence of ordered waypoints. 
Therefore, these sampling points are post-processed with a customized Travelling Salesman Problem (TSP)~\cite{LAPORTE1992231} solver to generate a shortest path but without returning to the starting point (by setting all edges that return to the starting point with 0 cost). We then route the robot along the path from its initial location to visit the remaining path waypoints.

\begin{algorithm}
    \caption{Information-Driven Planner}
    \label{algo:InfoPlanner}
        Given the desired number of waypoints $n$ \\
        \ForEach{$\bm{x} \in X$}{
            $V_1(\bm{x}) = I(Z_{\bm{x}}; Z_{X \setminus \{\bm{x}\}})$
        }
        \ForEach{\texttt{$i$ = 2 to $n$}} {
            \ForEach{$\bm{x} \in X$} { 
                initialize $V_i(\bm{x}) = -\infty$
            }
            \ForEach{$\bm{x}_{i-1} \in X$} { 
                \ForEach{$\bm{x}_{i} \in X \setminus \{\bm{x}_{1:i-1}\}$}{
                    $V_i(\bm{x}_{i}) = \max(I(Z_{\bm{x}_i}; Z_{X \setminus \{ \bm{x}_{1:i}\}}|Z_{\bm{x}_{1:i-1}}) + V_{i-1}(\bm{x}_{i-1}), V_i(\bm{x}_i))$
                }
            }
        }
        $\bm{x}_{n}^* = \argmax_{\bm{x}_{n} \in X}V_{n}(\bm{x}_{n})$ \\
        Backtrace to get $\bm{x}^* \triangleq (\bm{x}^*_{1}, \bm{x}^*_{2}, \dots , \bm{x}^*_{n})$ \\
\end{algorithm}


\subsection{Overall Framework} \label{sec:framework}

If the environment is dynamic, the prediction accuracy of GP degrades as time elapses because it does not incorporate the temporal variation of the environment. 
To address this issue, we re-estimate the hyperparameters at appropriate moments. The re-estimation triggering mechanism depends on two factors:
\begin{itemize}
\item The first factor stems from the computational concern. Since any re-estimation will be immediately followed by a re-planning of the future routing path, and because the computation time for the path planning is much more costly than that of the hyperparameter re-estimation. Thus, an appropriate frequency for the simultaneous re-estimation and re-planning needs to be determined to match the computational constraint. 
\item The second factor relates to the intensity of spatiotemporal variations. 
Since the kernel function that describes two points' spatial relation is an indicator of a GP's prediction capacity, 
thus the repetitive hyperparameter re-estimation of the kernel function should reflect the variation intensity of the environment. 
\end{itemize}

Note that, here we assume the temporal process is discrete. For example, the time can be discretized into a series of short intervals, durations, or horizons.
In our experiment, we slice the time into short intervals with equal length, 
and further assume that during each short interval, the environment is static, so that the environment can be regarded as  ``piecewise static" but all intervals form a dynamic process along the time dimension.
There can be many time steps in each interval and we assume that the robot performs a sampling operation at each time step, 
but only a small subset of samples are selected and used for hyperparameter optimization.

In our implementation, we use a measure, $\rho \in [0, 1]$, to decide the moment for triggering the re-estimation and re-planning processes. 
The measure $\rho$ represents the proportion of samples that are recently added to the current BV-set since last re-estimation.
The hyperparameter re-estimation and path re-planning are carried out if $\rho$ is above certain pre-defined threshold, $\rho_0$. 
Roughly, $\rho_0$ can be defined to be inversely proportional to the computational power and the intensity of environmental variation, and the higher the threshold, the less frequent the re-estimation. 

\begin{figure}
    \centering
    {\label{fig:SystemOverview}\includegraphics[height=2.5in]{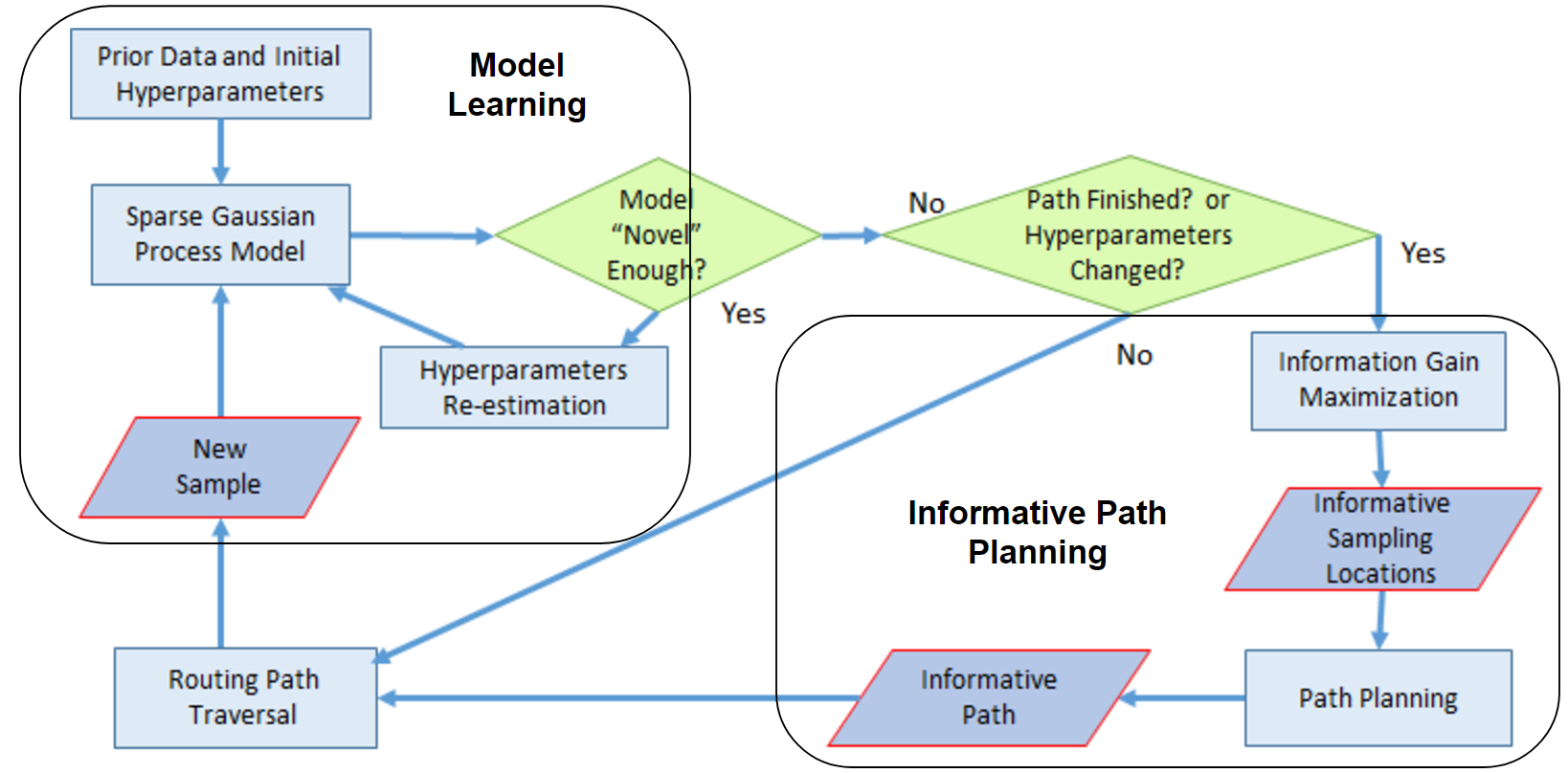}}
    \caption{
    Overview of the environmental monitoring system, which consists of two main components: environment model learning and informative path planning.
    }
    \label{fig:SystemOverview}
\end{figure}

The whole informative planning and online learning environmental monitoring framework is depicted in Fig.~\ref{fig:SystemOverview}, where the top-left part is the model learning component and the bottom right part is the informative planning component.
The vehicle's low level motion is handled in the {\em routing path traversal} block. 
We assume that while the vehicle is navigating from one waypoint to the next, the trajectory is roughly straight. 
The vehicle's motion under ocean current disturbance can be solved with different theoretical frameworks, for instance, 
it can be formulated under the control framework (e.g., see~\cite{mukhopadhyay2014collaborative}) or the decision-theoretic planning framework (e.g., see~\cite{ma2016information}). Due to limited space, we do not discuss low level motion in this paper.
The entire process is also pseudo-coded in Alg.~\ref{algo:onlinePlanning}.

\begin{algorithm} \label{algo:onlinePlanning}
    \caption{Online Learning and Informative Planning}
    Initialize SOGP \\
    \While{true} {
        $\rho = 0$ \quad /* for hyperparameter re-estimation */ \\
        Calculate sampling locations as described in \ref{sec:infoPlanner} \\
        Compute informative routing path, $P$, based on the sampling locations generated \\
        \ForEach{point $\bm{p} \in P$} {
            Do sampling on $\bm{p}$ to get a scalar value $v$\\
            Use $(\bm{p}, v)$ as a training point to update SOGP described in \ref{sec:OnlineGP} and \ref{sec:SparseGP} \\
            \If {$(\bm{p}, v)$ replaces some sample in the BV-set} {
                Increase $\rho$
            }
            \If {$\rho > \rho_0$ } {
                Do hyperparameter re-estimation described in \ref{sec:HyperTraining} \\
                break
            }
        }
    }
\end{algorithm}


\section{Simulation with Ocean Data}\label{sec:simulation}

\begin{figure}
    \centering
    \subfigure[]
    {\label{fig:truth}\includegraphics[height=1.6in]{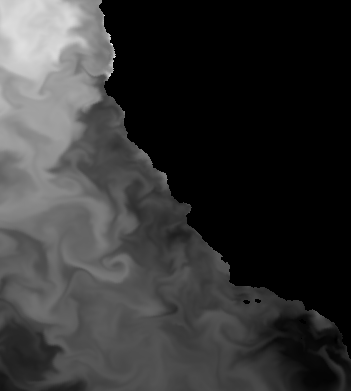}} \quad \quad
    \subfigure[]
    {\label{fig:gp_manual_49obs_annotated}\includegraphics[height=1.6in]{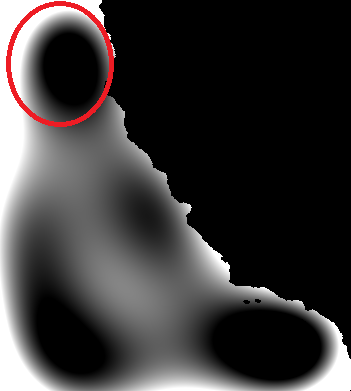}}
    \caption{(a) Salinity data obtained from ROMS. It is treated as a ground truth throughout the paper. The salinity value ranges from $31.97$ to $33.78$ (measurement unit: $\frac{g \text{ salt}}{kg \text{ sea water}}$) and it is remapped to grey scale values. Darker pixels indicate higher concentration of salinity; (b) The predicted model using GP without data-driven hyperparameter optimization.}
    \label{fig:HyperparametersAccu}
\end{figure}

\begin{figure}
    \centering
    \subfigure[]
    {\label{fig:DP_4_layer0_0_0obs_manual}\includegraphics[height=1.6in]{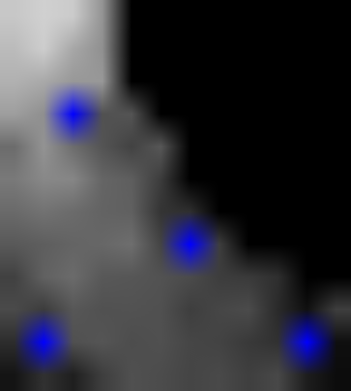}}
    \quad \quad
    \subfigure[]
    {\label{fig:DP_4_layer0_0_0obs_auto}\includegraphics[height=1.6in]{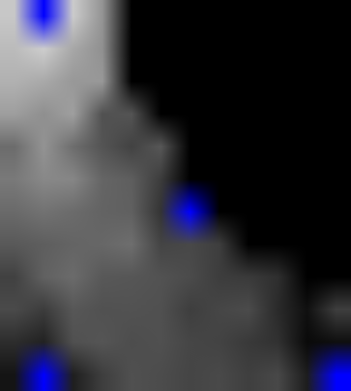}}
    \caption{Informative sampling locations before they are post-processed as paths. (a) Results using hyperparameters empirically set:
    $\{\sigma_n^2 = \exp(-2), \sigma_f^2 = \exp(2), l_x = \exp(1), l_y = \exp(1)\}$; 
    (b) Results using hyperparameters learned from data collected: 
    $\{\sigma_n^2 = \exp(-4.6), \sigma_f^2 = \exp(6.8), l_x = \exp(3.4), l_y = \exp(3.2)\}$.
    }
    \label{fig:HyperparametersWaypoints}
\end{figure}

\begin{figure*}
    \centering
    \subfigure[]
    {\label{fig:1robots_0observations_4DP_round4_lambda1_0}\includegraphics[height=1.1in]{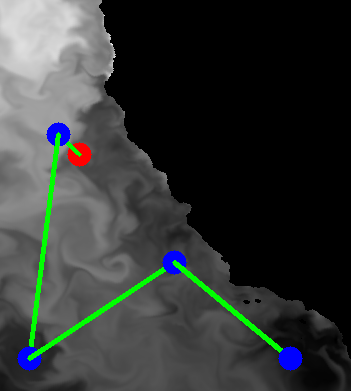}}
    \subfigure[]
    {\label{fig:1robots_100observations_4DP_round4_lambda1_1}\includegraphics[height=1.1in]{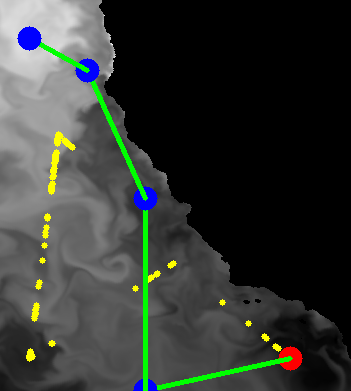}}
    \subfigure[]
    {\label{fig:1robots_100observations_4DP_round4_lambda1_2}\includegraphics[height=1.1in]{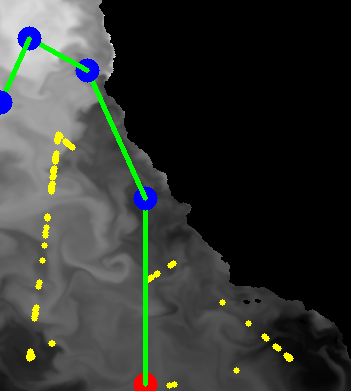}}
    \subfigure[]
    {\label{fig:1robots_100observations_4DP_round4_lambda1_3}\includegraphics[height=1.1in]{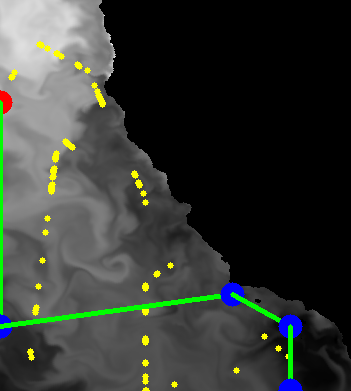}}
    \subfigure[]
    {\label{fig:1robots_0observations_4DP_round4_lambda1_4}\includegraphics[height=1.1in]{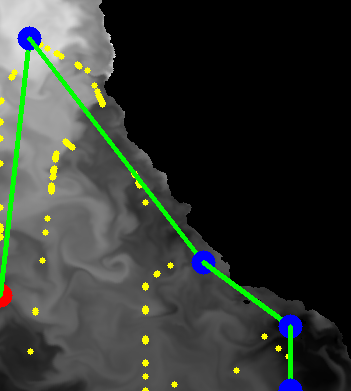}}
    \subfigure[]
    {\label{fig:1robots_100observations_4DP_round4_lambda1_5}\includegraphics[height=1.1in]{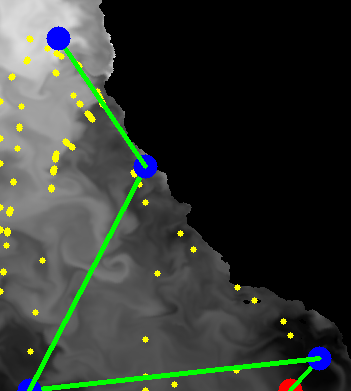}}
    \caption{(a)-(f) Informative paths resulted from subsequent re-plannings. The red and blue points represent the robot's starting locations and the informative sampling locations, respectively. The yellow dots denote the points stored in the SOGP BV-set.
    The latitude of the ocean area is ranged from $31.3^{\circ}$N to $43^{\circ}$N, and the longitude is ranged from $232.5^{\circ}$E to $243^{\circ}$E.
    The robot is initially deployed at (79, 236) in this example. Since the ocean is large and it requires many days to finish a traversal that can cover the area, we apply a much faster vehicle speed in simulation than that is used in reality. One may imagine that we simply re-scaled the space and time simultaneously in order to get a ``fast-forward play" of the results. 
}\label{fig:PathsMSE}
\end{figure*}

\begin{figure*}
    \centering
    \subfigure[]
    {\label{fig:sogp_0_stop1}\includegraphics[height=1.1in]{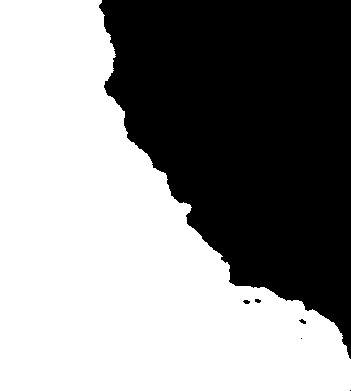}}
    \subfigure[]
    {\label{fig:sogp_510_stop0}\includegraphics[height=1.1in]{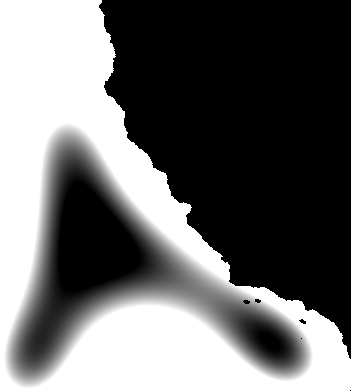}}
    \subfigure[]
    {\label{fig:sogp_663_stop1}\includegraphics[height=1.1in]{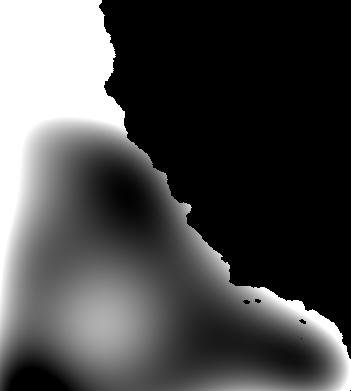}}
    \subfigure[]
    {\label{fig:sogp_1103_stop0}\includegraphics[height=1.1in]{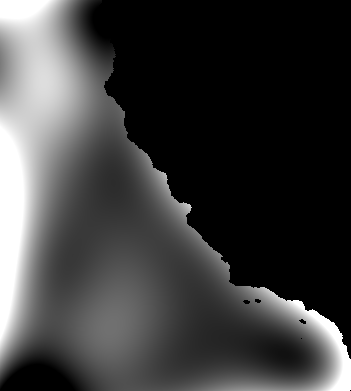}}
    \subfigure[]
    {\label{fig:sogp_1297_stop1}\includegraphics[height=1.1in]{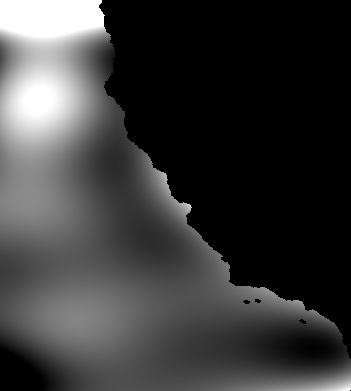}}
    \subfigure[]
    {\label{fig:sogp_1926_stop0}\includegraphics[height=1.1in]{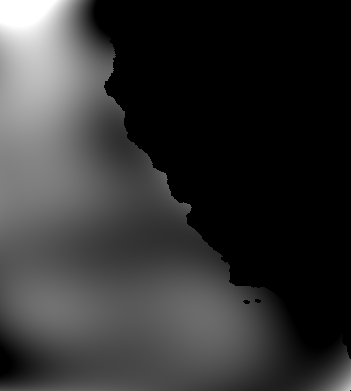}}
    \caption{(a)-(f) The learned salinity data map. Each corresponds to a step in Fig.~\ref{fig:PathsMSE}. 
    We assume the worst case scenario where no prior information is known, and this is well reflected from (a) where the entire estimated map is ``blank". After that, the accuracy of built map increased along with incoming data. }
     \label{fig:PredictedMap}
\end{figure*}

The proposed method is a generic framework which can be applied to any kinds of environments where the quantity of the interested phenomena varies spatially and temporally. That is, the environment can be terrestrial, aquatic or aerial and the dimension of the environment can be arbitrary too. Such examples include the concentration of pollutants in the air or the density of plankton in the ocean. Here, we validated our method in the scenario of ocean monitoring.
The simulation environment was constructed as a two dimensional ocean surface and we tessellated the environment into a grid map. In our experiments, we use ocean salinity data recently observed and post-processed by Regional Ocean Modeling System (ROMS)~\cite{shchepetkin_regional_2005}. 
Fig.~\ref{fig:truth} shows the post-processed salinity data in the Southern California Bight region. The data is represented as a scalar field (the black regions represent lands while the gray areas denote ocean), which is used as the ground truth for comparison.
We mapped the tessellated grip map with the ROMS data. Specifically, the resolution of the grid map is $351\times391$ pixels, and after the mapping each pixel corresponds to around $2.85\times2.85km^2$ of the real ocean dimension.
In addition, each pixel is also labeled with ocean coordinate information (latitude and longitude), and the environmental attributes such as surface salinity value, as well as the time stamp of the measurement.  
When the vehicle traverses a pixel, a sampling operation is performed and the salinity value is retrieved. 
Since the component of computing the informative sampling locations is more expensive, 
a down-sampling of ROMS data to a desired resolution is performed to alleviate the computational cost. 
In our experiments, the resolution for the sampling points generation (for path planning) is down-sampled to $12 \times 12$.
We implemented a sparse online variant of GP (SOGP) built upon the open-source GP library {\em libgp}~\cite{blum2013optimization}.

First, we investigate the model prediction accuracy using un-tuned hyperparameters, i.e., hyperparameter values are set empirically/manually instead of data-driven. 
Fig.~\ref{fig:gp_manual_49obs_annotated} shows a prediction result with 50 prior random samples and manually set hyperparameters $\bm{\theta}_0 = \{\sigma_n^2 = \exp(-2), \sigma_f^2 = \exp(2), l_x = \exp(1), l_y = \exp(1)\}$. 
We can observe that the prediction does not match well with the ground truth (see the area circled in red). 
This is because that, as mentioned in \ref{sec:gp} the hyperparameters of a GP ``control" the spatial relation between two independent measurements (data points in the field), 
and a particular set of samples need a particular configuration of hyperparameters for the best regression. 
Thus, empirically setting values is very likely to be suboptimal.
In addition to that, we have also investigated and compared the generated informative sampling points (locations) using empirical and data-driven hyperparameters. 
Here the data-driven method produces optimized hyperparameters  ${\hat{\bm\theta}}=\{\sigma_n^2 = \exp(-4.6), \sigma_f^2 = \exp(6.8), l_x = \exp(3.4), l_y = \exp(3.2)\}$, from which we can observe that every single parameter of the optimized set has changed in order to fit the data.
Fig.~\ref{fig:DP_4_layer0_0_0obs_manual} and \ref{fig:DP_4_layer0_0_0obs_auto} show generated informative sampling points from  the environmental model with 
manually-set and data-driven hyperparameters, respectively.
We can see that the relative distances between points (and the covered areas) in Fig.~\ref{fig:DP_4_layer0_0_0obs_auto} are larger than those in Fig.~\ref{fig:DP_4_layer0_0_0obs_manual}. This is mainly affected by $\bm{l}$, which controls the pairwise spatial correlations. 
Intuitively, larger values of $\bm{l}$ imply larger range of similarity in the vicinity. 
Formally, given a sampled point, we have a larger area of confidence in the vicinity, hence we should reach further to explore the uncertain area.
That is why the generated sampling points in Fig.~\ref{fig:DP_4_layer0_0_0obs_auto} are more spread out than in Fig.~\ref{fig:DP_4_layer0_0_0obs_manual}. 

Therefore, we use the data-driven planning and learning method to update the hyperparameters online.
The process is demonstrated in Fig.~\ref{fig:PathsMSE}, where a total number of 
$2000$ sampling operations have been performed. 
Fig.~\ref{fig:1robots_0observations_4DP_round4_lambda1_0} to~\ref{fig:1robots_100observations_4DP_round4_lambda1_5} illustrate a series of snapshots at re-planning moments, so that all subsequent newly generated path segments can be fully seen.
Specifically, a new informative path is computed if a previous path has been completed or the GP hyperparameters have been re-estimated and updated. 
For example, Figs.~\ref{fig:1robots_100observations_4DP_round4_lambda1_1}, \ref{fig:1robots_100observations_4DP_round4_lambda1_3}, and \ref{fig:1robots_100observations_4DP_round4_lambda1_5} are the ones after completion of previous paths,  whereas the remaining sub-figures are the ones generated after hyperparameter re-estimations.
The red and blue points stand for the robot's current starting position and the informative sampling locations, respectively; the yellow dots represent the points stored in the SOGP BV-set. The robot launched from a random shore location $(79, 236)$ 
and performed the sampling operations at each time step along the planned path. The initial hyperparameters are set to $\{\sigma_n^2 = \exp(-2), \sigma_f^2 = \exp(2), l_x = \exp(1), l_y = \exp(1)\}$ and we emulated the memory limit by setting the maximum size of the BV-set as $m = 100$. The threshold is set as $\rho_0 = 0.6$. 
The distribution patterns of the yellow dots in Fig.~\ref{fig:1robots_0observations_4DP_round4_lambda1_0} to~\ref{fig:1robots_100observations_4DP_round4_lambda1_5} reveal the sparseness of BV-set, indicating that as the robot gradually explores the whole map, the BV-set only stores those points that are the most useful for predicting the model. 
The corresponding predicted maps are shown in Fig.~\ref{fig:PredictedMap}, from which we can see that the constructed models constantly converge to the ground truth and are able to characterize the general patterns of the environment at the final stages.

\begin{figure*}
    \centering
    \subfigure[]
    {\label{fig:thin_1robots_0observations_4LawnMower_round4_lambda1_0}\includegraphics[height=1.1in]{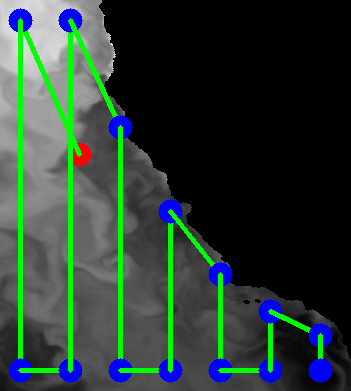}}
    \subfigure[]
    {\label{fig:thin_1robots_0observations_4LawnMower_round4_lambda1_1}\includegraphics[height=1.1in]{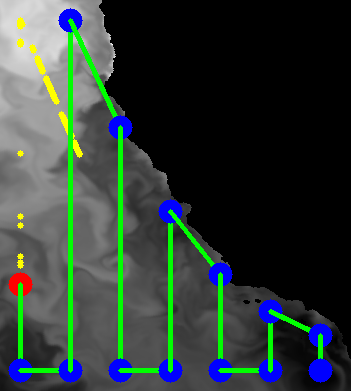}}
    \subfigure[]
    {\label{fig:thin_1robots_0observations_4LawnMower_round4_lambda1_2}\includegraphics[height=1.1in]{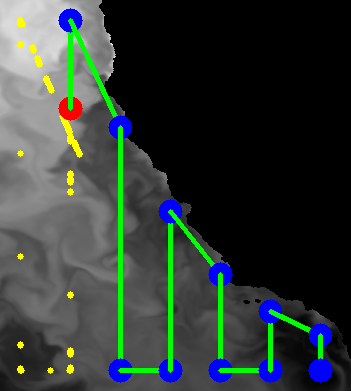}}
    \subfigure[]
    {\label{fig:thin_1robots_0observations_4LawnMower_round4_lambda1_3}\includegraphics[height=1.1in]{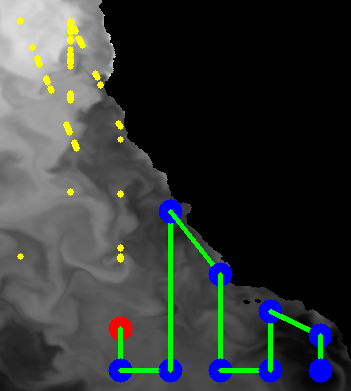}}
    \subfigure[]
    {\label{fig:thin_1robots_0observations_4LawnMower_round4_lambda1_4}\includegraphics[height=1.1in]{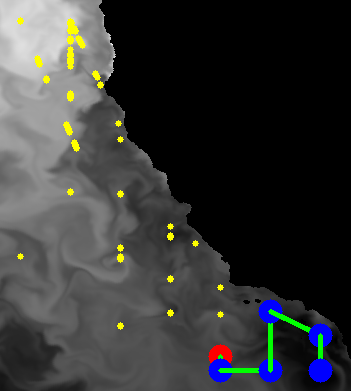}}
    \subfigure[]
    {\label{fig:thin_1robots_0observations_4LawnMower_round4_lambda1_5}\includegraphics[height=1.1in]{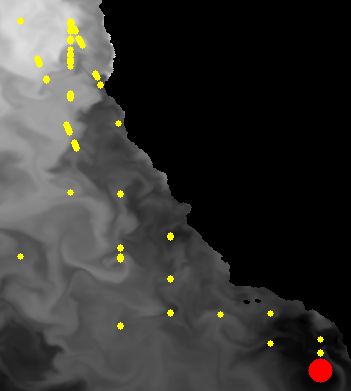}}
    \caption{
    Lawnmower sampling locations are manually set beforehand. This figure shows a lawnmower pattern with relatively high sweeping resolution. It takes about $1812$ sampling operations to finish the whole path.
}\label{fig:thin_LawnPathsMSE}
\end{figure*}

\begin{figure*}
    \centering
    \subfigure[]
    {\label{fig:thin_lawn_sogp_0_stop1}\includegraphics[height=1.1in]{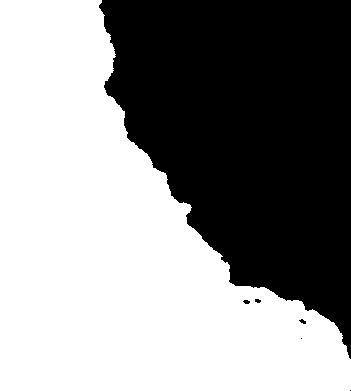}}
    \subfigure[]
    {\label{fig:thin_lawn_sogp_400_stop0}\includegraphics[height=1.1in]{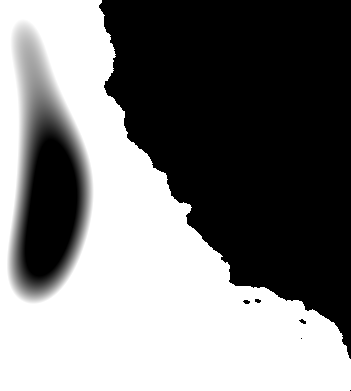}}
    \subfigure[]
    {\label{fig:thin_lawn_sogp_800_stop0}\includegraphics[height=1.1in]{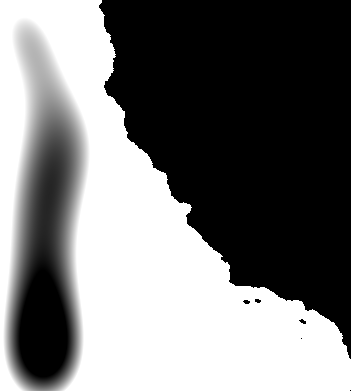}}
    \subfigure[]
    {\label{fig:thin_lawn_sogp_1200_stop0}\includegraphics[height=1.1in]{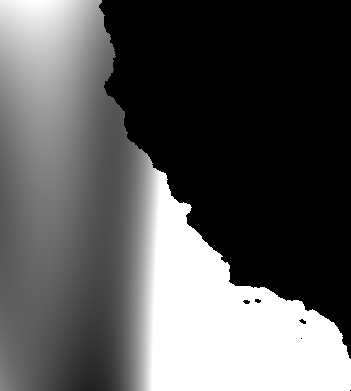}}
    \subfigure[]
    {\label{fig:thin_lawn_sogp_1600_stop0}\includegraphics[height=1.1in]{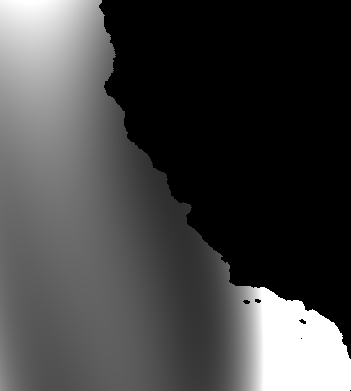}}
    \subfigure[]
    {\label{fig:thin_lawn_sogp_1812_stop0}\includegraphics[height=1.1in]{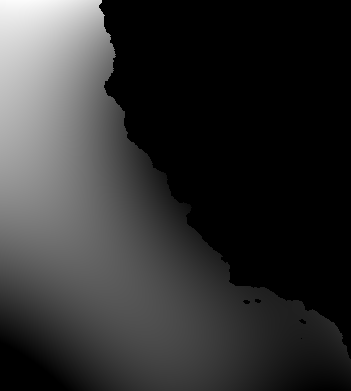}}
    \caption{
    (a)-(f) The learned salinity data maps from the high-resolution lawnmower sampling approach. Each map corresponds to a step in Fig.~\ref{fig:thin_LawnPathsMSE}.
    }
    \label{fig:thin_LawnPredictedMap}
\end{figure*}

\begin{figure*}
    \centering
    \subfigure[]
    {\label{fig:wide_1robots_0observations_4LawnMower_round4_lambda1_0}\includegraphics[height=1.1in]{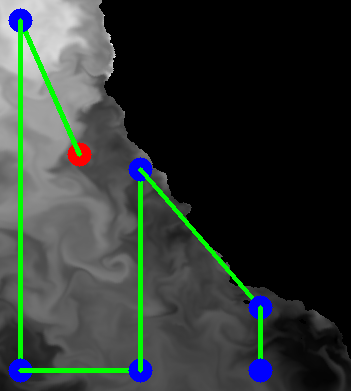}}
    \subfigure[]
    {\label{fig:wide_1robots_0observations_4LawnMower_round4_lambda1_1}\includegraphics[height=1.1in]{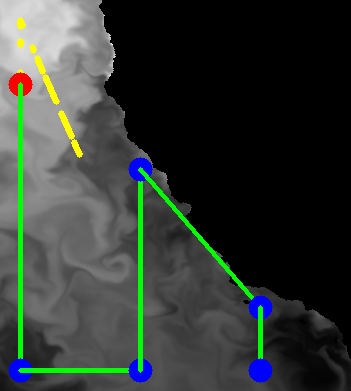}}
    \subfigure[]
    {\label{fig:wide_1robots_0observations_4LawnMower_round4_lambda1_2}\includegraphics[height=1.1in]{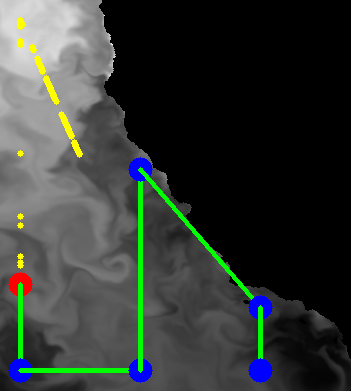}}
    \subfigure[]
    {\label{fig:wide_1robots_0observations_4LawnMower_round4_lambda1_3}\includegraphics[height=1.1in]{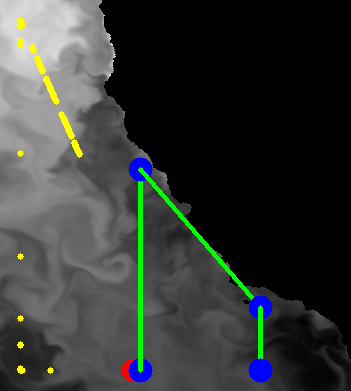}}
    \subfigure[]
    {\label{fig:wide_1robots_0observations_4LawnMower_round4_lambda1_4}\includegraphics[height=1.1in]{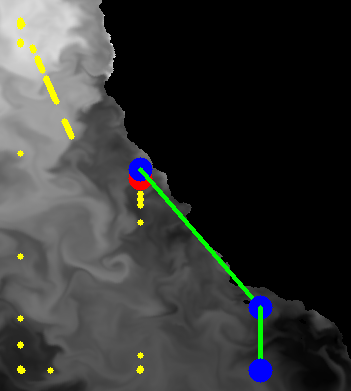}}
    \subfigure[]
    {\label{fig:wide_1robots_0observations_4LawnMower_round4_lambda1_5}\includegraphics[height=1.1in]{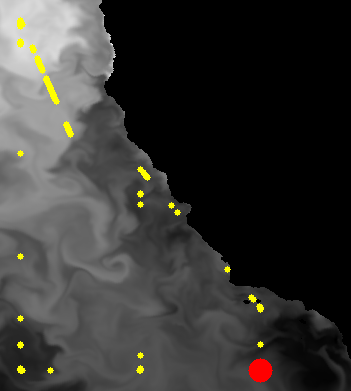}}
    \caption{
    Lawnmower sampling path with a lower sweeping resolution. It takes about $1000$ sampling operations to finish the whole path.
}\label{fig:wide_LawnPathsMSE}
\end{figure*}

\begin{figure*}
    \centering
    \subfigure[]
    {\label{fig:wide_lawn_sogp_0_stop1}\includegraphics[height=1.1in]{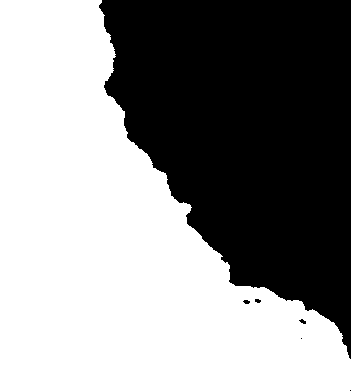}}
    \subfigure[]
    {\label{fig:wide_lawn_sogp_200_stop0}\includegraphics[height=1.1in]{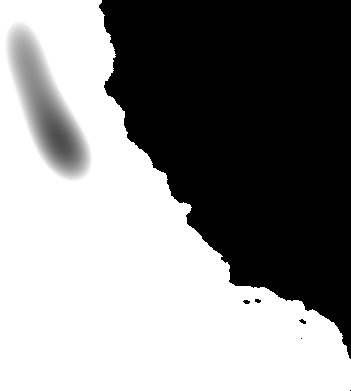}}
    \subfigure[]
    {\label{fig:wide_lawn_sogp_400_stop0}\includegraphics[height=1.1in]{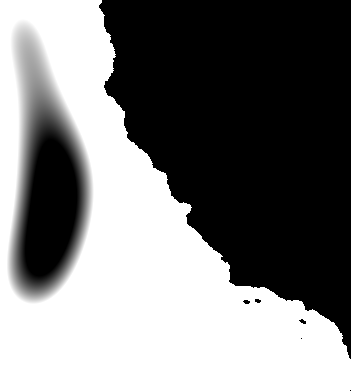}}
    \subfigure[]
    {\label{fig:wide_lawn_sogp_600_stop0}\includegraphics[height=1.1in]{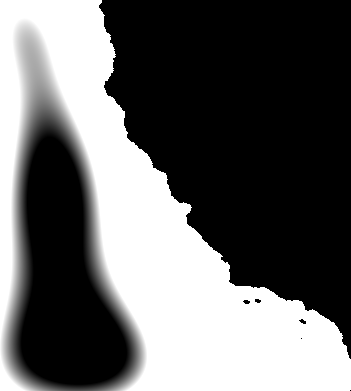}}
    \subfigure[]
    {\label{fig:wide_lawn_sogp_800_stop0}\includegraphics[height=1.1in]{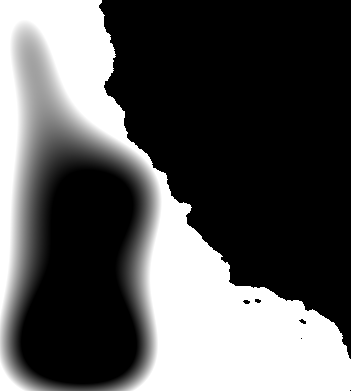}}
    \subfigure[]
    {\label{fig:wide_lawn_sogp_993_stop0}\includegraphics[height=1.1in]{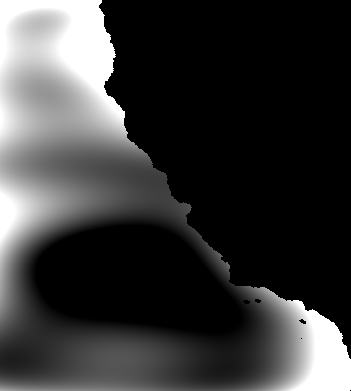}}
    \caption{
    (a)-(f) The learned salinity data maps from the low-resolution lawnmower sampling approach. Each map corresponds to a step in Fig.~\ref{fig:wide_LawnPathsMSE}.
    }
    \label{fig:wide_LawnPredictedMap}
\end{figure*}

To evaluate the model prediction accuracy, 
we have compared our method with two other non-informative frameworks: a lawnmower sampling approach which has been widely used for estimating static environments, and a Monte Carlo random sampling approach where the area can be explored and covered quickly.
Note that the only difference between our method and other control experimental methods is how the waypoints are generated. 
The other routines such as the online learning and the hyperparameter re-estimation remain the same since for all sampling and planning methods we need to learn the hyperparameters and reduce computational cost by using sparse samples.

\begin{figure}
    \centering
    \subfigure[]
    {\label{fig:79236_jfr}\includegraphics[height=2in]{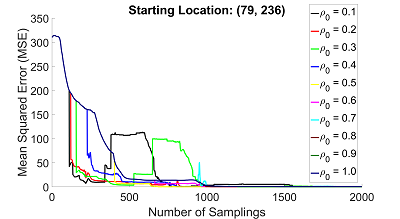}}
    \subfigure[]
    {\label{fig:info_random_lawn}\includegraphics[height=2in]{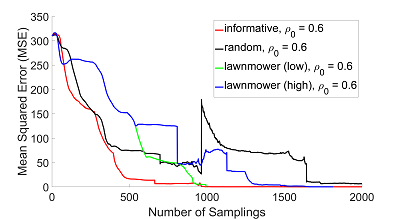}}
    \subfigure[]
    {\label{fig:info_random_lawn_zoom}\includegraphics[height=2in]{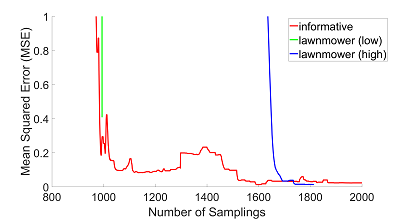}}
    \caption{(a) The MSE results from multiple trials  with threshold $\rho$ tuned in a range $\{0.1, 0.2, \dots, 1.0\}$. The $y$-axis is the MSE value while the $x$-axis is the number of sampling operations. 
    (b) Statistics of all mentioned sampling strategies with the same threshold $\rho_0 = 0.6$. 
    (c) Partially enlarged plot of (b) for a closer comparison.
    }
    \label{fig:mse}
\end{figure}

We first investigate the lawnmower approach.
Fig.~\ref{fig:thin_LawnPathsMSE} and Fig.~\ref{fig:wide_LawnPathsMSE} show the paths generated in lawnmower patterns.
The difference between them is that the spacing of sweeps of Fig.~\ref{fig:thin_LawnPathsMSE} is narrower than that of Fig.~\ref{fig:wide_LawnPathsMSE}. Such different sweeping ``resolutions" result in different amounts of time or distance to finish sweeping the area. 
Fig.~\ref{fig:thin_LawnPredictedMap} and Fig.~\ref{fig:wide_LawnPredictedMap} are the obtained predicted maps, respectively.
The prediction errors of all methods (compared with the ground truth) are shown in Fig.~\ref{fig:mse}, from which we can see that our proposed method achieves to reduce to an error at around 0.2 with approximately 1000 sampling steps.
To best compare with the lawnmower approach, we first set the prediction error threshold as 0.2, i.e., the lawnmower needs to reduce the prediction error to 0.2 before it terminates the sampling.
For the scenario in Fig.~\ref{fig:thin_LawnPathsMSE} with high sweeping resolution, we found that it requires about $1700$ sampling operations to complete, indicating almost twice time (distance) in order to achieve the same map accuracy. An even higher resolution of lawnmower sweeping requires even longer sampling time and farther traveling distance.
Note that in this approach, there is no re-planning involved because the entire path is pre-planned and fixed. 
It is as expected that high resolution lawnmower sweeping can eventually obtain a better prediction result given sufficient time since it samples and evaluates at a finer level.  
This also implies that the lawnmower approach can be best used in estimating static environments, but may be slow and insensitive in dynamic environments.

In addition, we also look into the lawnmower scheme with fixed number of sampling operations.
Specifically, since the informative sampling method needs around 1000 sampling operations to achieve an error of 0.2, 
thus we set the total sampling operations of the lawnmower approach to be 1000 as well. 
In this case, the spacing of lawnmower sweeps can be calculated.
Fig.~\ref{fig:wide_LawnPathsMSE} is the resultant lawnmower pattern, and we can see that the sweeping resolution is much lower than that of Fig.~\ref{fig:thin_LawnPathsMSE}.
The performance in terms of map prediction errors can be found in Fig.~\ref{fig:info_random_lawn}, where the result labeled {\em lawnmower (low)} in green is from this scenario. We can see that our informative method reduces the prediction error much faster.

\begin{figure*}
    \centering
    \subfigure[]
    {\label{fig:1robots_0observations_4Random_round4_lambda1_0}\includegraphics[height=1.1in]{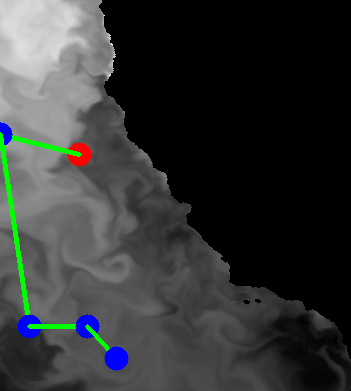}}
    \subfigure[]
    {\label{fig:1robots_0observations_4Random_round4_lambda1_1}\includegraphics[height=1.1in]{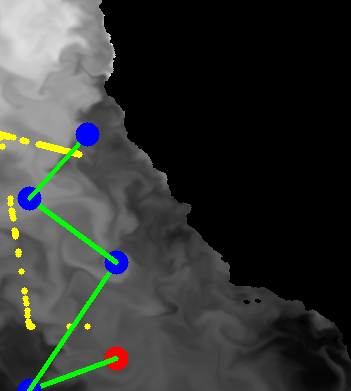}}
    \subfigure[]
    {\label{fig:1robots_100observations_4Random_round4_lambda1_2}\includegraphics[height=1.1in]{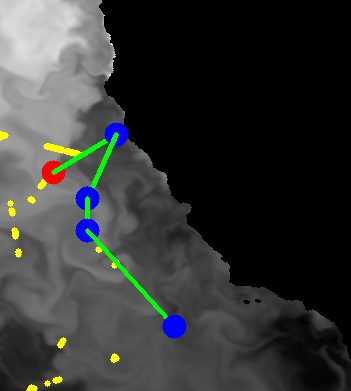}}
    \subfigure[]
    {\label{fig:1robots_100observations_4Random_round4_lambda1_3}\includegraphics[height=1.1in]{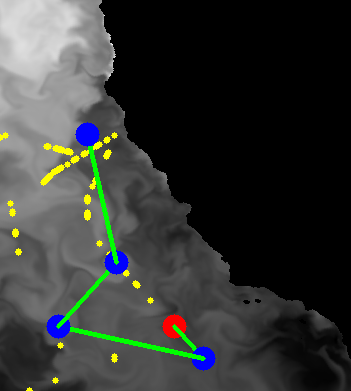}}
    \subfigure[]
    {\label{fig:1robots_100observations_4Random_round4_lambda1_6}\includegraphics[height=1.1in]{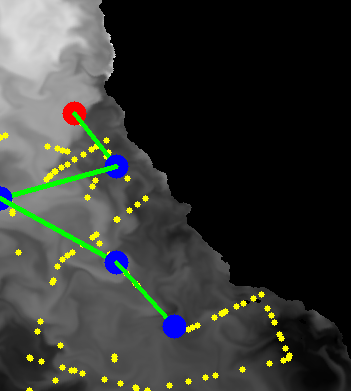}}
    \subfigure[]
    {\label{fig:1robots_100observations_4Random_round4_lambda1_7}\includegraphics[height=1.1in]{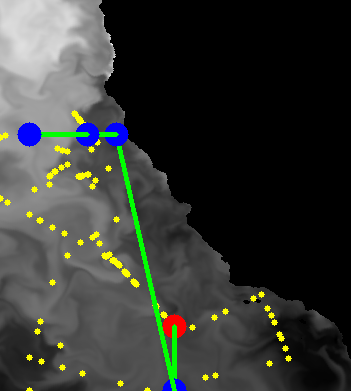}}
    \caption{
    Sampling locations are generated using a Monte Carlo random sampling strategy, so that the space is expected to be explored more quickly than the lawnmower approach. Other routines such as the online learning and hyperparameter re-estimation are the same as those of Fig.~\ref{fig:PathsMSE}.
}\label{fig:RandomPathsMSE}
\end{figure*}

\begin{figure*}
    \centering
    \subfigure[]
    {\label{fig:random_sogp_0_stop1}\includegraphics[height=1.1in]{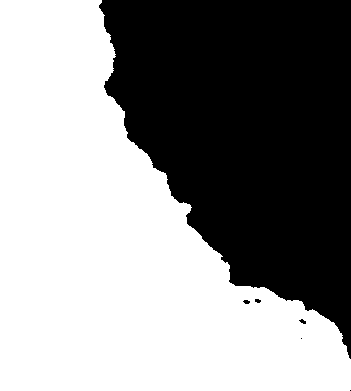}}
    \subfigure[]
    {\label{fig:random_sogp_365_stop0}\includegraphics[height=1.1in]{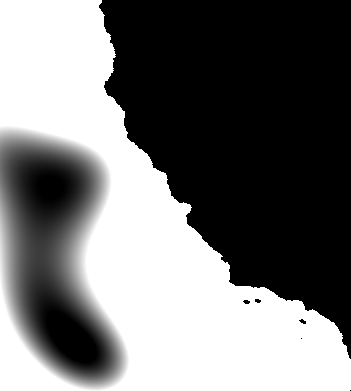}}
    \subfigure[]
    {\label{fig:random_sogp_697_stop1}\includegraphics[height=1.1in]{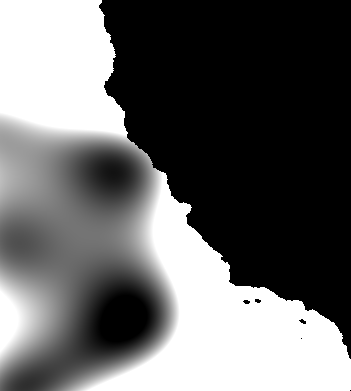}}
    \subfigure[]
    {\label{fig:random_sogp_956_stop0}\includegraphics[height=1.1in]{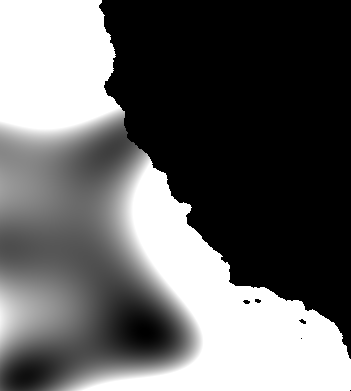}}
    \subfigure[]
    {\label{fig:random_sogp_1642_stop1}\includegraphics[height=1.1in]{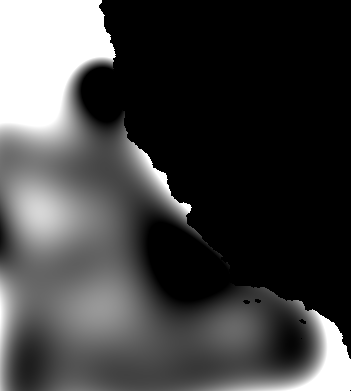}}
    \subfigure[]
    {\label{fig:random_sogp_1995_stop0}\includegraphics[height=1.1in]{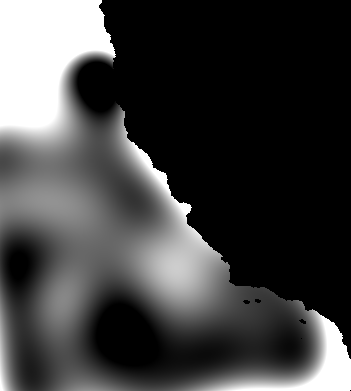}}
    \caption{(a)-(h) The learned salinity data maps from the Monte Carlo random sampling. Each map corresponds to a step in Fig.~\ref{fig:RandomPathsMSE}. 
    Note that certain local regions (e.g., the top area) are not explored and reached by this method.
    }
    \label{fig:RandomPredictedMap}
\end{figure*}

Next, we compare with the Monte Carlo random sampling approach since this simple strategy can quickly explore and cover the space in a uniform way.  
We also finish the same number of sampling operations used in the informative sampling method.
The resultant sampling locations/paths and the retained samples from the random sampling approach are shown in Fig.~\ref{fig:RandomPathsMSE};
the corresponding predicted environmental models are shown in Fig.~\ref{fig:RandomPredictedMap}.
Among all those snapshots, Fig.~\ref{fig:random_sogp_365_stop0}, \ref{fig:random_sogp_956_stop0}, and \ref{fig:random_sogp_1995_stop0} are the ones generated after finishing previous paths, and the remaining ones are triggered after the updates of hyperparameters.  
Since the random sampling scheme does not take into account of informativeness, 
we can see that compared to the informative approach, the generated waypoints are less spread out for exploring those most unknown regions. 
The final predicted data map (model) is shown in Fig.~\ref{fig:random_sogp_1995_stop0}, which clearly reveals that there is still a mis-prediction in the top area where it has never been explored at all.


We then examine the prediction errors between the predicted data maps (i.e., GP's predicted mean values) and the ground truth, as shown in Fig.~\ref{fig:mse}.
First, Fig.~\ref{fig:79236_jfr} shows statistics of the proposed method under different thresholds $\rho_0$.
The $x$-axis represents the number of sampling operations, which is roughly proportional to the travel time (or distance). 
The $y$-axis is the MSE calculated using the whole map as a testing set. 
The figure reveals that, in general every setting follows a descending trend with the error being constantly reduced along the exploration process. 
By adjusting thresholds $\rho_0$, we can see that there are more error fluctuations for low $\rho_0$ values. A possible reason is that, if the explored regions do not yet well cover the environment, the hyperparameter re-estimation might optimize only among some local regions rather than the entire map, causing a loss of generality and an overfitting problem. 
However, since the actual relation between two independent data points in the environment may change over time, a low $\rho_0$ value has the advantage of adapting to the changes more quickly because lower $\rho_0$ indicates that the hyperparameter re-estimations are done more frequently.
Detailed comparisons among differing sampling schemes are shown in Fig.~\ref{fig:info_random_lawn}, where the tail (or ending) part of statistics are enlarged in Fig.~\ref{fig:info_random_lawn_zoom}.
By comparing these different curves, we can see that our method decreases the prediction error at the fastest rate since the slope is steeper than others (the error is reduced to $0.2$ with around $1000$ samples).
The two lawnmower approaches are the slowest, with the low-resolution sweeping slightly faster.
The Monte Carlo random method lies in the middle, but becomes unstable after it reaches certain level of prediction accuracy.
For the two implementations of lawnmower sampling, although the low-resolution version can reduce the error at a faster rate since it covers the space more quickly (the error is reduced to 0.4 with around 1000 samples), with sufficient long time the high-resolution version can achieve higher accuracy (the error is reduced to below 0.2 with around 1700 samples). Again, the large time cost is a main factor that prevents the lawnmower sampling approach from being used in the (highly dynamic) spatiotemporal environment estimation problem.  


\begin{figure*}
    \centering
    \subfigure[]
    {\label{fig:var_2}\includegraphics[height=0.75in]{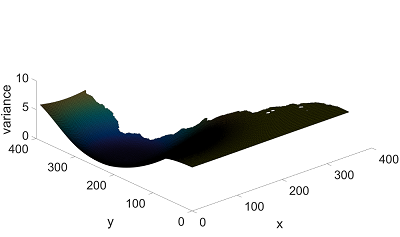}} \ \ 
    \subfigure[]
    {\label{fig:var_3}\includegraphics[height=0.75in]{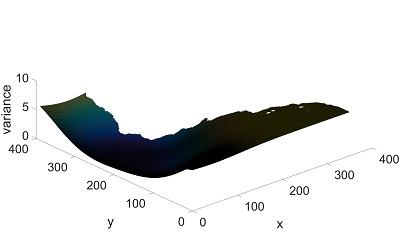}}\ \ 
    \subfigure[]
    {\label{fig:var_4}\includegraphics[height=0.75in]{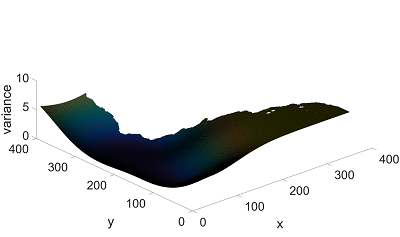}}\ \ 
    \subfigure[]
    {\label{fig:var_5}\includegraphics[height=0.75in]{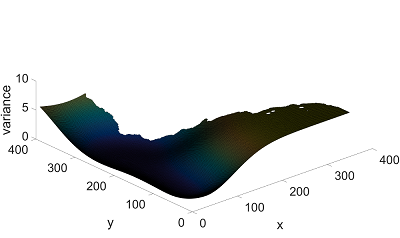}}\ \ 
    \subfigure[]
    {\label{fig:var_6}\includegraphics[height=0.75in]{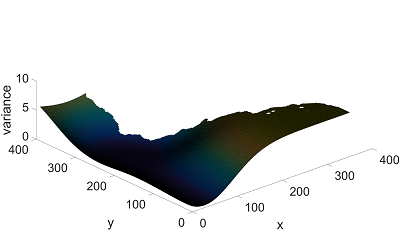}}\ \ 
    \subfigure[]
    {\label{fig:var_7}\includegraphics[height=0.75in]{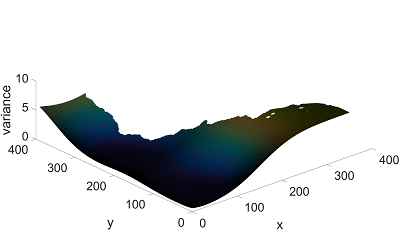}}\ \ 
    \subfigure[]
    {\label{fig:var_8}\includegraphics[height=0.75in]{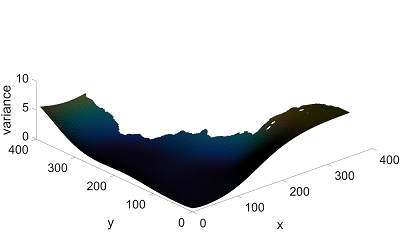}}\ \ 
    \subfigure[]
    {\label{fig:var_9}\includegraphics[height=0.75in]{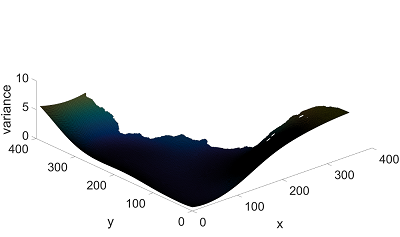}}\ \ 
    \subfigure[]
    {\label{fig:var_10}\includegraphics[height=0.75in]{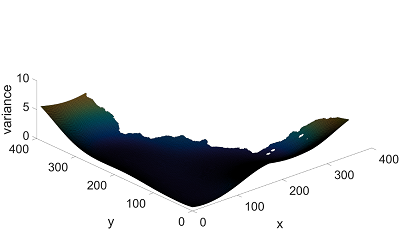}}\ \ 
    \subfigure[]
    {\label{fig:var_last}\includegraphics[height=0.75in]{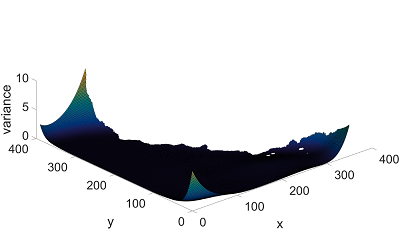}}
    \caption{ Maps of prediction variances from the proposed method. (a)-(i) Variances reduce as the robot follows planned path and collects data samples. (j) The final variance map that corresponds to the moments in Fig.~\ref{fig:1robots_100observations_4DP_round4_lambda1_5} and Fig.~\ref{fig:sogp_1926_stop0}.}
    \label{fig:VarMap}
\end{figure*}

\begin{figure*}
    \centering
    {\label{fig:random_variance_1995}\includegraphics[height=0.75in]{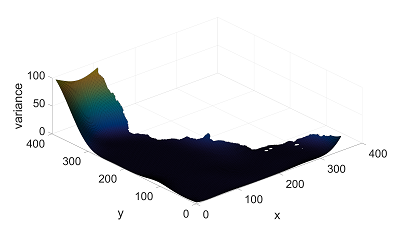}}
    \caption{The final variance map of the random sampling approach. It corresponds to the moment in Fig.~\ref{fig:1robots_100observations_4Random_round4_lambda1_7} and Fig.~\ref{fig:random_sogp_1995_stop0}.}
    \label{fig:RandomVarMap}
\end{figure*}

We also assess the model prediction variances. 
We create a variance map on which each ``pixel" records the prediction variance of that point.  
Fig.~\ref{fig:VarMap} illustrates a series of variance maps along the sampling operations.
We can see that the map gradually ``falls towards the ground", indicating a decrease of prediction variances along the robot's exploration.  The variance map in the last time step is depicted in Fig.~\ref{fig:var_last}. 
Compared to Fig.~\ref{fig:random_variance_1995}, which is also the final stage variance map of the random approach, we can conclude that our proposed method succeeds in exploring most of the unknown and uncertain regions while the random scheme fails to do so.

\begin{figure}
    \centering
    \subfigure[]
    {\label{fig:frame_count0_truth}\includegraphics[width=1.2in]{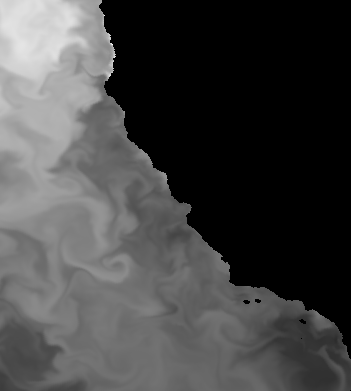}}
    \subfigure[]
    {\label{fig:frame_count2_truth}\includegraphics[width=1.2in]{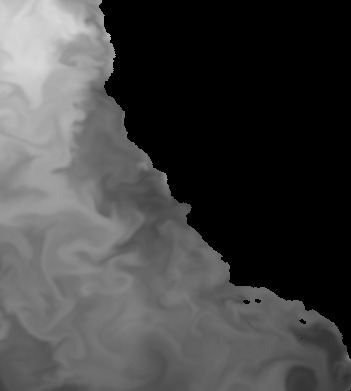}}
    \subfigure[]
    {\label{fig:frame_count4_truth}\includegraphics[width=1.2in]{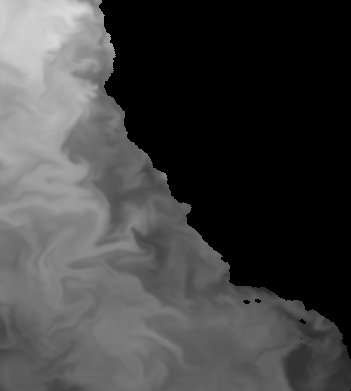}}
    \subfigure[]
    {\label{fig:frame_count6_truth}\includegraphics[width=1.2in]{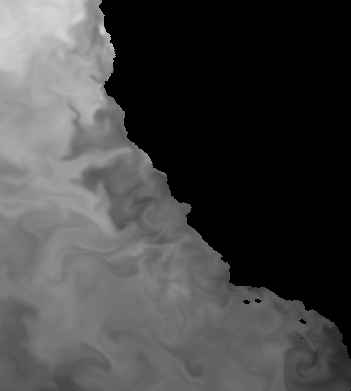}}
    \subfigure[]
    {\label{fig:frame_count8_truth}\includegraphics[width=1.2in]{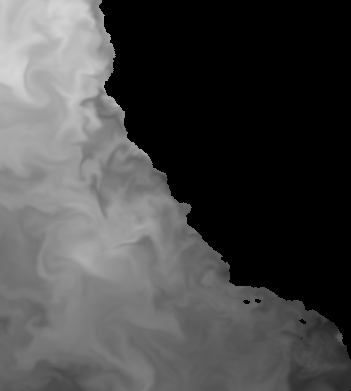}}
    \caption{The spatiotemporal environment is constructed by concatenating a sequence of groundtruth salinity maps at different times. }
    \label{fig:dynamic_environment}
\end{figure}

\begin{figure}
    \centering
    \subfigure[]
    {\label{fig:groundtruth_variation}\includegraphics[height=2in]{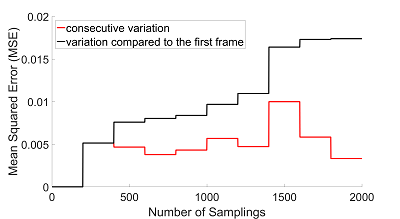}}
    \subfigure[]
    {\label{fig:dynamic_mse}\includegraphics[height=2in]{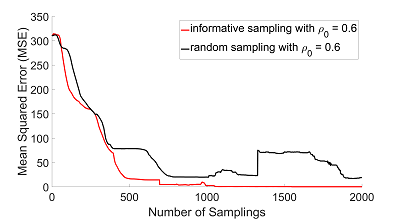}}
    \subfigure[]
    {\label{fig:mse_zoom}\includegraphics[height=2in]{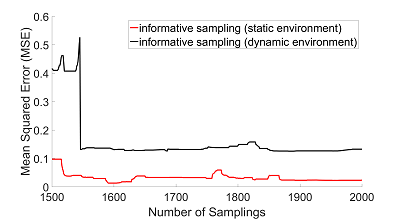}}
    \caption{(a) The salinity variation measured in MSE. The red curve is calculated by comparing each current frame with its previous one, while the black one is by comparing each current frame with the first frame; (b) MSE comparison between the informative sampling method and the random sampling scheme in a dynamic environment; (c) MSE comparison between the  static and dynamic environments, for the informative sampling method upon convergence. }
    \label{fig:dynamic}
\end{figure}

Lastly, we investigate the scenario of dynamic environments, where there are variations along the temporal process. 
To construct such a spatiotemporal environment, we sequentially concatenate a series of ``static environment frames". 
One may also imagine that we slice the time into a series of short intervals (horizons) with equal length and assume that during each small fixed-duration time interval, the environment is static. Therefore, the environment can be regarded as  ``piecewise static" but all intervals form a dynamic process along the time dimension.
In greater detail,
we change the underlying environment frames every $200$ sampling operations. 
The variations of different salinity maps are shown in Fig.~\ref{fig:dynamic_environment}. 
In Fig.~\ref{fig:groundtruth_variation}, we express the temporal variation by calculating the MSE between each frame/interval and the previous frame (red line) or the first frame (black line). 
Since the environment does not change within each interval, the curves are thus piecewise constant functions. 

We compare between the proposed framework and the random sampling scheme in the dynamic environment. 
The results are shown in Fig.~\ref{fig:dynamic_mse}, from which we can see that 
the error of our approach can still converge to a small value whereas there is a large error after $2000$ sampling operations for the random sampling approach.
The large error in the random sampling approach is due to the addition of temporal dynamics on top of the spatial variations.
More specifically, a time-varying environment  brings in more uncertainty than the static one.
Since GP provides a measure of modelling uncertainty, or equivalently the informativeness, and since our informative planning framework can best acquire the informativeness, the collected data of which is then used to update the latent hyperparameters that support the GP, 
therefore, our method can well capture the spatiotemporal variations, leading to large error reduction.
In contrast, the random sampling method does not take into account any uncertainty and so its exploration contains no informativeness. And this is why the random sampling strategy fails.   

It is worth mentioning that, although our approach can quickly converge, the final error is larger than its counterpart application in the static environment. 
To show this, we zoom in the plot by focusing on the sampling operations from $1500$ to $2000$, as shown in Fig.~\ref{fig:mse_zoom}. 
We can see that the error of the dynamic case is slightly higher than the static one by a value of $0.1$.
This makes sense as the time-varying environment has larger uncertainty than the static one. Note that there's a sudden decrease of error at $1545$, it's again because of the hyperparameter re-estimation.


\section{Field Trials with an Autonomous Surface Vehicle}

\begin{figure}[t]
    \centering
    \subfigure[]
    {\label{fig:uscasv}\includegraphics[height=1.7in]{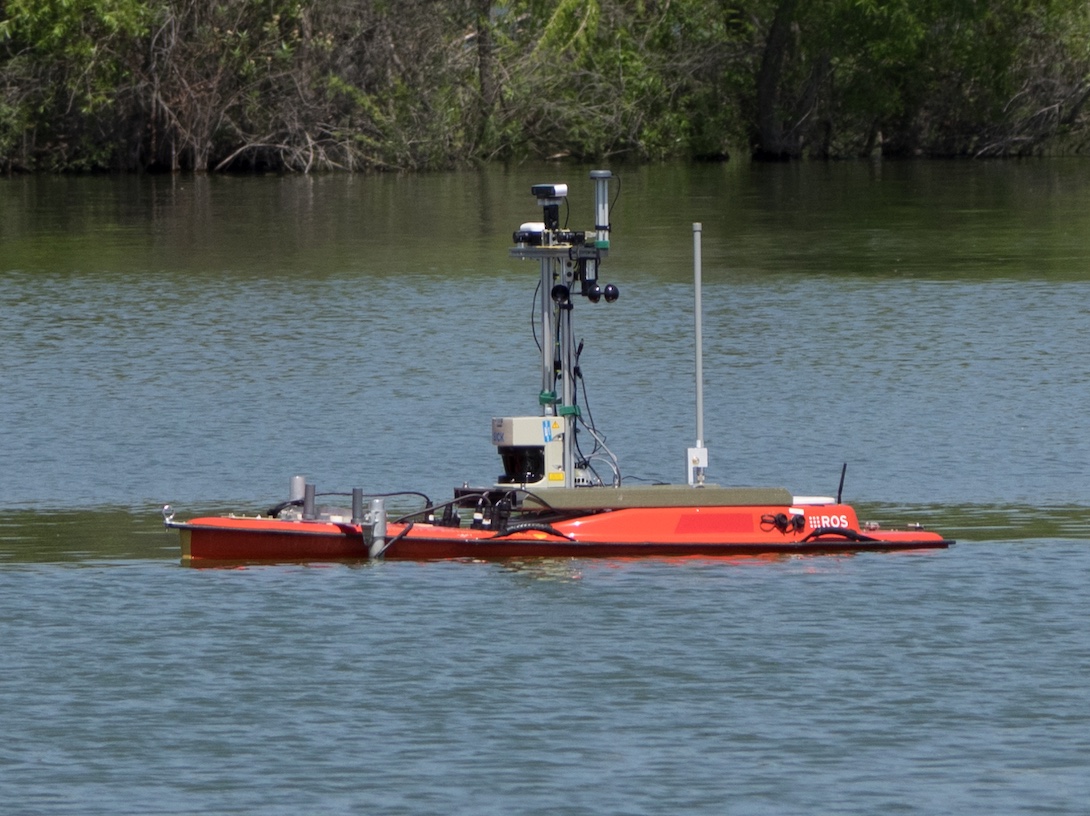}}
    \subfigure[]
    {\label{fig:interface}\includegraphics[height=1.7in]{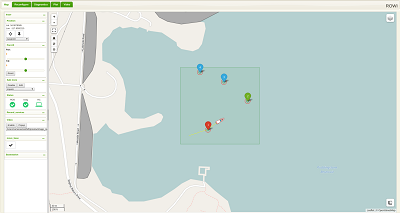}}
    \caption{(a) The ASV developed at the University of Southern California; (b) The planning interface of the ASV. The bounding box pre-defines a restricted mission area and the numbered pinpoints are the waypoints that the ASV is going to traverse.}
\end{figure}

Field trials were conducted with an autonomous surface vehicle (ASV) developed at the University of Southern California (USC), pictured in Fig.~\ref{fig:uscasv}, at Puddingstone Reservoir in San Dimas, California. The USC ASV is around $2\ m$ long and $0.8\ m$ wide, actuated by two electrical thrusters at the back and a rudder and is therefore underactuated as it can not perform any sway motions. 
It is equipped with an IMU and Real-Time Kinematic (RTK) GPS for localization, an acoustic depth sensor to measure depth and onboard computers for control. The depth sensor used in these trials was a Furuno~235DST-PWE depth sensor which has a range of $100\ m$, beam width of $7^{\circ}$, sampling rate of $1$~Hz and resolution of $0.1$~m. In addition it has sensors, such as forward looking sonar, laser range finder and stereo cameras, that were not used in this trial.

In the field trial, we focus on sampling in the static environment. 
There are two reasons for doing so.
The first reason lies in the ease of obtaining ground truth, with which we can perform comparisons and evaluations. 
In contrast, in a time-varying environment this is almost impossible unless we use many vehicles simultaneously where those vehicles are mainly used for constructing ground-truth. 
The second reason lies in that, as we analyzed in the simulation section, a time-varying environment can be 
represented by a sequence of static environments. One can regard the static environment as a static interval of a long-term time-varying environment.

The field trial site is shown in Fig.~\ref{fig:interface}. 
Note that most inland lakes and reservoirs contain not only little temporal variation but also little spatial variation, 
for most environment attributes such as temperature, salinity, pollution or nutrition contents, etc. These physical or chemical attributes are easily spread along with the dynamic water flow and they are typically uniform across the entire environment.
Therefore, we opt to model the {\em water depth} from the water surface to the bottom of the lake. 
The water depth gives us enough spatial variation and 
it stays constant during and between runs so we can easily use previously collected ground truth for comparison. 

\begin{figure*}[t] 
    \centering
    \subfigure[]
    {\label{fig:ground1}\includegraphics[height=2.3in]{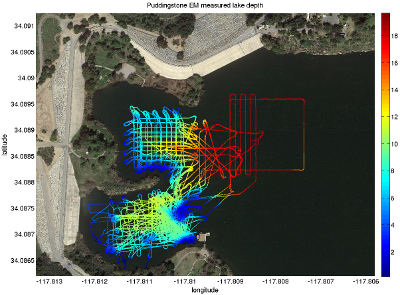}}
    \subfigure[]
    {\label{fig:ground2}\includegraphics[height=2.3in]{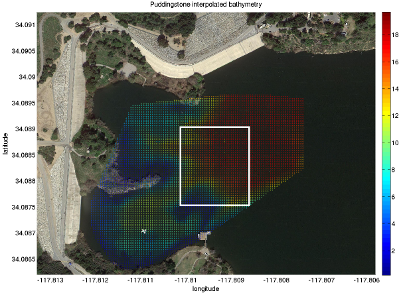}}
    \subfigure[]
    {\label{fig:lake_groundtruth}\includegraphics[height=2in]{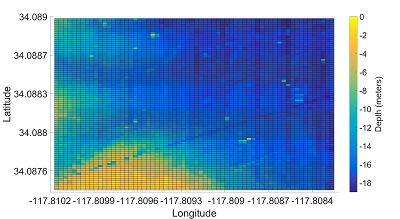}}
    \subfigure[]
    {\label{fig:lake_groundtruth_gray}\includegraphics[height=2in]{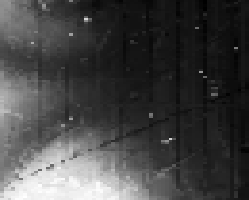}}
    \caption{(a) Lawnmower paths for estimating the grounth truth; (b) Constructed ground truth map. The $x$-axis is the longitude while the $y$-axis is the latitude. The depth unit is in meters; 
    (c) A restricted bounding-box area is used for our field trial. This area corresponds to the white box in Fig.(b) (d) The same groundtruth area in gray-scale. The darker, the deeper.}
    \label{fig:LakerGroundTruth}
\end{figure*}

The ground truth was obtained by combining water depth data from an acoustic depth sensor from multiple prior runs of our YSI EcoMapper AUV, as shown in Fig.~\ref{fig:ground1}. We collected the depth data using the comprehensive lawnmower approach, resulting in a dense depth map for the experimental area. The recorded depth data was combined to form a grid based depth map using linear interpolation, as shown in Fig.~\ref{fig:ground2}. In these figures, the color tone corresponds to the water depth where warmer color represents deeper depth while colder color represents shallower depth. The maximum depth is roughly $19$ meters.

\begin{figure}[t]
    \centering
    \subfigure[]
    {\label{fig:lake_random_path}\includegraphics[height=1.7in]{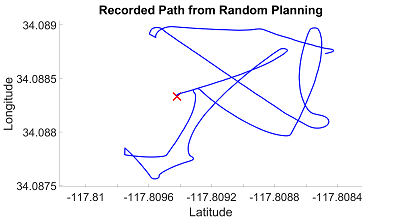}}
    \subfigure[]
    {\label{fig:lake_dp_path}\includegraphics[height=1.7in]{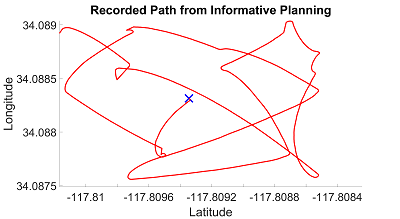}}
    \caption{Recorded paths in the field trial. The `x' is the starting location. (a) The paths from the random sampling method;  (b) The paths from this proposed method.}
    \label{fig:lake_path}
\end{figure}

In our experiment, in order to shorten the duration of each trial and exclude the possibility of bumping into obstacles, we intentionally define the experimental area to be a sub-region of the groundtruth. It is set to be a square ranging from latitude $34.0875$ to $34.0890$ and longitude $-117.8082$ to $-117.8102$. The size is about $170 \times 165$ squared meters. The boat is restricted to run within this predefined area for about 20 minutes per one trial. The initial starting location is set to be roughly at the midpoint of the area. The initial starting location can actually be arbitrary since we assume no prior samples in advance. That is, we have no idea about the depth at all before we start collecting the data. However if we have some prior samples (e.g. collected from previous trials or from some static sensors), we might tend to set the initial starting location to be away from the area where prior samples are located. In this way, some traveling cost can be saved by not passing the known area.

Next, we describe how the ASV is controlled. 
Due to the underactuated nature of the vehicle, following a pre-planned trajectory can be futile when large disturbances are present, such as strong winds. 
We therefore do high level control of the ASV by specifying a sequence of waypoints for it to traverse. The waypoints, each of which consists of a latitude and longitude coordinate and a ``hit radius", are visited in the defined order and are regarded as having been reached and completed once the ASV is within the waypoint's hit radius. To reach any subsequent waypoint, the controller does not pre-plan an accurate trajectory, instead it points the ASV towards this local target at all times and maintains the commanded velocity. 
This means that when there is no disturbance, the vehicle's actual trajectory is roughly a straight line between two consecutive waypoints; otherwise,  the vehicle may drift sideways due to external perturbations.

The ASV localization is performed using an Extended Kalman Filter that combines output from an onboard IMU and a Real-Time Kinematic (RTK) GPS receiver. We run an RTK basestation on shore that broadcasts correction data to the ASV through the WiFi communication link. By filtering with RTK the localization precision is within $10$~cm.

Fig.~\ref{fig:interface} shows the interface of our ASV and the example of controlling it through waypoints specification. The bounding box is exactly the predefined area described in the previous paragraph, and the pinpoints are the waypoints that the ASV is going to traversed. The number on the pinpoints denotes the traversal sequence. The sampling rate of the depth pinger is $1$~Hz, which is the same as the one used for collecting the ground truth data. We also set the speed of the ASV to be a constant of $1.5\ m/s$ throughout the entire trial. Because the ASV is always running in a constant speed (excluding the effect of the wind), we can assume that the distance traveled or the number of sampling operations are equivalent to the meaning of duration. The maximum size of the BV-set and the threshold are also set to $m = 100$ and $\rho_0 = 0.6$, as in the simulation with ocean data. The initial hyperparameters of the covariance function are empirically set to $\{\sigma_n^2 = \exp(-2), \sigma_f^2 = \exp(4), l_x = \exp(-8), l_y = \exp(-8)\}$. Here the $\bm{l}$ hyperparameters are significantly smaller than the ones in Sec.\ref{sec:simulation}. It is because we are now interested in the depth structure in the reservoir rather than the salinity in the ocean. It is the different properties between the two interested targets that cause different settings of hyperparameters.


In the field trials, we also compare our approach with the random approach described in Sec.\ref{sec:simulation}. First we compare the recorded paths resulted from the random approach and our approach in Fig.~\ref{fig:lake_path}. The $\times$ denotes the starting location of the boat, which is roughly at the center of the whole area. Fig.~\ref{fig:lake_random_path} shows the paths resulted from uniformly selecting random waypoints for the boat to traverse while Fig.~\ref{fig:lake_dp_path} shows the result of our proposed approach. Given the same constant speed of the boat and the time, they both collected the same amount of data and travelled the same distance. One might notice the boat is wandering at some certain points, this is due to the boat is recalculating a new path since it has either completed the current batch of waypoints or the hyperparameter re-estimation procedure has been triggered. During recalculation, we intentionally let the ASV park and disengage its motors at the current location and stop sampling. Because there's wind blowing the ASV away from the parking location in the field and the ASV would try to go back to the parking location, as a result, it seems wandering around the parking location. In Fig.~\ref{fig:lake_random_path}, we can see there are some areas missed to be explored by the boat, for example, the right lower and left parts, while in Fig.~\ref{fig:lake_dp_path}, most of the area is already covered in the same amount of time. The results shows that our approach is better in terms of coverage given the same amount of time. 

\begin{figure}[t]
    \centering
    \subfigure[]
    {\label{fig:random_depth_1128}\includegraphics[height=2in]{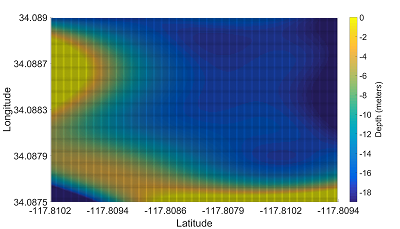}}
    \subfigure[]
    {\label{fig:random_depth_1128_gray}\includegraphics[height=2in]{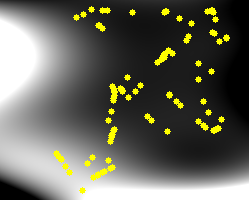}}
    \caption{(a) The depth estimate result from the random sampling approach; (b) The same depth estimate in gray-scale. The yellow dots denote the data points stored in the SOGP BV-set.}
    \label{fig:random_lake_depth}
\end{figure}

Next we compare the performance in terms of modeling the environment. Fig.~\ref{fig:random_lake_depth} shows the final depth map estimated from the SOGP using samples collected from the random paths. Fig.~\ref{fig:random_depth_1128_gray} also shows the locations of preserved samples stored in the SOGP BV-set. We can see that in the areas missed by the boat, the depth estimate is inaccurate because few samples in the missed areas are stored. However for the area where samples are stored, the SOGP can captures the depth structure fairly well, for example, the top-right part is relatively deep while the bottom-left part is shallow.

\begin{figure}[t]
    \centering
    \subfigure[]
    {\label{fig:dp_depth_0002}\includegraphics[height=0.7in]{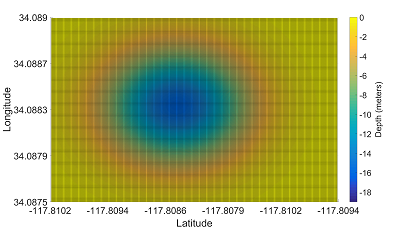}}
        \subfigure[]
    {\label{fig:dp_depth_0152}\includegraphics[height=0.7in]{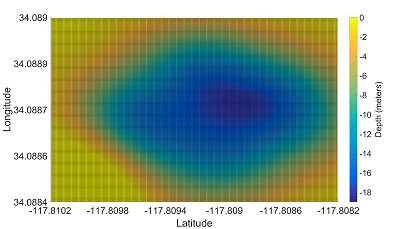}}
        \subfigure[]
    {\label{fig:dp_depth_0302}\includegraphics[height=0.7in]{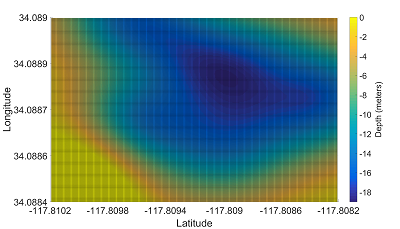}}
        \subfigure[]
    {\label{fig:dp_depth_0452}\includegraphics[height=0.7in]{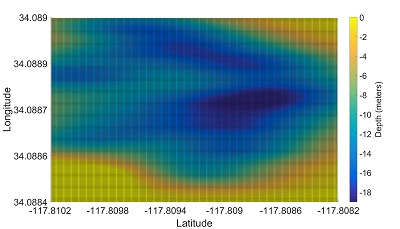}}
        \subfigure[]
    {\label{fig:dp_depth_0602}\includegraphics[height=0.7in]{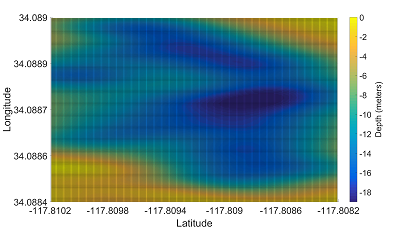}}
        \subfigure[]
    {\label{fig:dp_depth_0752}\includegraphics[height=0.7in]{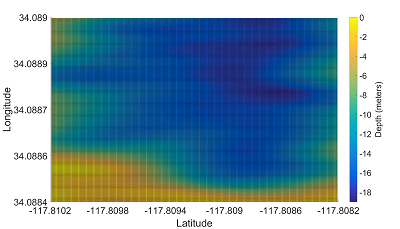}}
        \subfigure[]
    {\label{fig:dp_depth_0902}\includegraphics[height=0.7in]{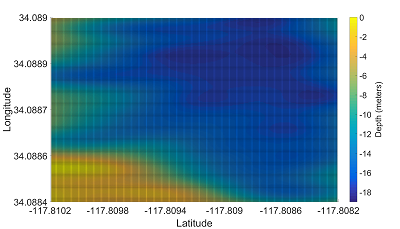}}
        \subfigure[]
    {\label{fig:dp_depth_1052}\includegraphics[height=0.7in]{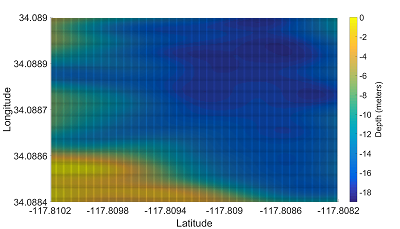}}
    \subfigure[]
    {\label{fig:dp_depth_1128}\includegraphics[height=0.7in]{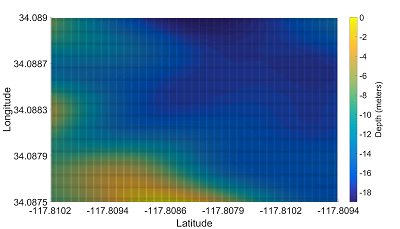}}
    \subfigure[]
    {\label{fig:dp_depth_1128_gray}\includegraphics[height=2in]{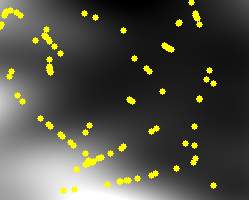}}
    \caption{(a)-(i) The progress of depth-map estimation along the informative sampling paths; (j) The final estimated depth map, which also corresponds to state (i). The retained data points (yellow dots) have a good coverage of the environment. }
    \label{fig:lake_depth}
\end{figure}

Fig.~\ref{fig:lake_depth} shows the progress of the estimated depth map along the informative paths proposed by us. The duration between two consecutive depth maps is $150$ sampling operations and Fig.~\ref{fig:dp_depth_1128} is the final result.
Initially we have no idea about the field, the depth in most of the region are estimated as $0$, but as it progresses, we are getting better understanding about the overall depth structure. From the final result, we can see that it captures the distribution of depth a lot better than the random approach. It successfully models the three shallow regions on the left and the deep region on the right. From Fig.~\ref{fig:dp_depth_1128_gray}, we can also see that the samples stored in the BV-set are distributed to describe the depth structure accordingly.

\begin{figure*}[t] 
    \centering
    {\label{fig:lake_mse}\includegraphics[height=2in]{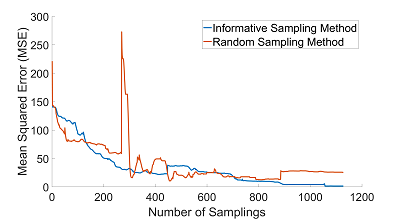}}
    \caption{The MSE comparison between the informative sampling approach and the random sampling method in the field trial. The $y$-axis is the MSE value while the $x$-axis is the number of sampling operations.}
    \label{fig:LakeMSE}
\end{figure*}

Lastly, the corresponding numerical estimation errors are also compared. Fig.~\ref{fig:LakeMSE} shows the MSE plot of the two approaches calculated from comparing with the ground truth data. The $y$-axis is the MSE value while the $x$-axis is the total number of sampling operations. The overall error in both approaches are decreasing as time goes although there are some significant fluctuations and peaks (e.g. there is a peak at $x = 300$ in random approach). Again, it is possibly because the regions are not yet well covered, the hyperparameter re-estimation procedure might optimize only among some local regions rather than the entire map. An obvious difference between the two approaches is that the final error of our approach has already converged down to $1.4035$ in the end while the other is still relatively high. This indicates that the informative planning in our approach can significantly reduce the cost of exploration and we believe that with larger areas defined, the reduction should be more remarkable.

\section{Conclusions}

Environmental monitoring entails persistent presence by robots. This
suggests that both planning and learning are likely to constitute
critical components of any robotic system built for monitoring. In
this paper, we present an informative planning and online learning
method that enables an autonomous aquatic vehicle to effectively
perform persistent ocean monitoring tasks. Our proposed framework
iterates between a planning component that is designed to collect data
with the maximal information content, and a sparse Gaussian Process
learning component where the environmental data map and hyperparameters are learned online by
selecting and utilizing only a subset of data that makes the greatest
contribution. 
We have conducted both simulations and field trials by comparing with a non-informative sampling method, 
and the results show that our method produces a good match between the predicted models and the
ground truths, with superior decrements for both prediction errors and map variances.  

In the future, we are interested in learning dynamic environmental data maps with continuous temporal models.
Specifically, the Gaussian Process provides a basis essentially used for building spatial models, 
and our presented work has extended the spatial model to a spatiotemporal process by incorporating a series of 
sliced temporal intervals which can be regarded piecewise static.
Such a framework can very well catch up with the environmental dynamics by taking advantage of sparse samples with the most contribution.
However, 
since the sliced temporal process is discrete and relies on the resolution of each time interval, thus may fail in 
capturing the temporal changes of the environment at a very fine level.
Therefore, our near future work will be investigating solutions of integrating continuous temporal models or processes.

\subsubsection*{Acknowledgments}
This work was supported in part by grants from ONR (N00014-14-1-0536), NSF (IIS-1619319 ) and USDA/NIFA (2017-67007-26154).
The authors would like to thank Stephanie Kemna for her help on collecting some of field data.

{ 
\bibliographystyle{apalike}
\bibliography{reference.bib}

\begin{thebibliography}{}

\bibitem[Anderson and Gaston, 2013]{anderson2013lightweight}
Anderson, K. and Gaston, K.~J. (2013).
\newblock Lightweight unmanned aerial vehicles will revolutionize spatial
  ecology.
\newblock {\em Frontiers in Ecology and the Environment}, 11(3):138--146.

\bibitem[Azuma et~al., 2012]{AzSaSePa2012}
Azuma, S., Sakar, M.~S., and Pappas, G. (2012).
\newblock Stochastic source seeking by mobile robots.
\newblock {\em Automatic Control, IEEE Transactions on}, 57(9):2308--2321.

\bibitem[Bai et~al., 2016]{bai2016}
Bai, S., Wang, J., Chen, F., and Englot, B. (2016).
\newblock Information-theoretic exploration with bayesian optimization.
\newblock In {\em 2016 {IEEE/RSJ} International Conference on Intelligent
  Robots and Systems, {IROS} 2016, Daejeon, South Korea, October 9-14, 2016},
  pages 1816--1822.

\bibitem[Binney et~al., 2010]{Binney-2010-642}
Binney, J., Krause, A., and Sukhatme, G.~S. (2010).
\newblock Informative path planning for an autonomous underwater vehicle.
\newblock In {\em International Conference on Robotics and Automation}, pages
  4791--4796.

\bibitem[Binney et~al., 2013]{binney13}
Binney, J., Krause, A., and Sukhatme, G.~S. (2013).
\newblock Optimizing waypoints for monitoring spatiotemporal phenomena.
\newblock {\em International Journal on Robotics Research (IJRR)},
  32(8):873--888.

\bibitem[Blum and Riedmiller, 2013]{blum2013optimization}
Blum, M. and Riedmiller, M.~A. (2013).
\newblock Optimization of gaussian process hyperparameters using rprop.
\newblock In {\em ESANN}.

\bibitem[Brooks et~al., 2008]{brooks2008gaussian}
Brooks, A., Makarenko, A., and Upcroft, B. (2008).
\newblock Gaussian process models for indoor and outdoor sensor-centric robot
  localization.
\newblock {\em IEEE Transactions on Robotics}, 24(6):1341--1351.

\bibitem[Cao et~al., 2013]{Cao2013}
Cao, N., Low, K.~H., and Dolan, J.~M. (2013).
\newblock Multi-robot informative path planning for active sensing of
  environmental phenomena: A tale of two algorithms.
\newblock In {\em Proceedings of the 2013 International Conference on
  Autonomous Agents and Multi-agent Systems}, pages 7--14.

\bibitem[Chang et~al., 2013]{ChWeWeWeZh2013}
Chang, D., Wu, W., Webster, D., Weissburg, M., and Zhang, F. (2013).
\newblock A bio-inspired plume tracking algorithm for mobile sensing swarms in
  turbulent flow.
\newblock In {\em Robotics and Automation (ICRA), 2013 IEEE International
  Conference on}, pages 921--926.

\bibitem[Cortes, 2007]{Co2007}
Cortes, J. (2007).
\newblock Distributed gradient ascent of random fields by robotic sensor
  networks.
\newblock In {\em Decision and Control, 2007 46th IEEE Conference on}, pages
  3120--3126.

\bibitem[Cristianini et~al., 2001]{ello2001kernel}
Cristianini, N., Elisseeff, A., Shawe-Taylor, J., and Kandola, J. (2001).
\newblock On kernel-target alignment.
\newblock {\em Advances in neural information processing systems}.

\bibitem[Csat{\'o} and Opper, 2002]{csato2002sparse}
Csat{\'o}, L. and Opper, M. (2002).
\newblock Sparse on-line gaussian processes.
\newblock {\em Neural computation}, 14(3):641--668.

\bibitem[Dunbabin and Marques, 2012]{Dunbabin12}
Dunbabin, M. and Marques, L. (2012).
\newblock Robotics for environmental monitoring: Significant advancements and
  applications.
\newblock {\em IEEE Robot. Autom. Mag.}, 19(1):24 -- 39.

\bibitem[Erickson et~al., 2015]{multi-sensor2015}
Erickson, P., Cline, M., Tirpankar, N., and Henderson, T. (2015).
\newblock Gaussian processes for multi-sensor environmental monitoring.
\newblock In {\em 2015 IEEE International Conference on Multisensor Fusion and
  Integration for Intelligent Systems (MFI)}, pages 208--213.

\bibitem[Fackler and Haight, 2014]{fackler2014monitoring}
Fackler, P.~L. and Haight, R.~G. (2014).
\newblock Monitoring as a partially observable decision problem.
\newblock {\em Resource and Energy Economics}, 37:226--241.

\bibitem[Farrell et~al., 2005]{farrell2005chemical}
Farrell, J.~A., Pang, S., and Li, W. (2005).
\newblock Chemical plume tracing via an autonomous underwater vehicle.
\newblock {\em IEEE Journal of Oceanic Engineering}, 30(2):428--442.

\bibitem[Fiorelli et~al., 2003]{Fiorelli03adaptivesampling}
Fiorelli, E., Bhatta, P., and Leonard, N.~E. (2003).
\newblock Adaptive sampling using feedback control of an autonomous underwater
  glider fleet.
\newblock In {\em Proc. 13th Int. Symposium on Unmanned Untethered Submersible
  Tech}, pages 1--16.

\bibitem[Guestrin et~al., 2005]{guestrin2005near}
Guestrin, C., Krause, A., and Singh, A.~P. (2005).
\newblock Near-optimal sensor placements in gaussian processes.
\newblock In {\em Proceedings of the 22nd international conference on Machine
  learning}, pages 265--272. ACM.

\bibitem[Hajieghrary et~al., 2015]{hajieghrary2015information}
Hajieghrary, H., Tom, A.~F., Hsieh, M.~A., et~al. (2015).
\newblock An information theoretic source seeking strategy for plume tracking
  in 3d turbulent fields.
\newblock In {\em 2015 IEEE International Symposium on Safety, Security, and
  Rescue Robotics (SSRR)}, pages 1--8. IEEE.

\bibitem[Hengl, 2009]{hengl2009practical}
Hengl, T. (2009).
\newblock {\em A practical guide to geostatistical mapping}, volume~52.
\newblock Hengl.

\bibitem[Hollinger et~al., 2016]{hollinger2016learning}
Hollinger, G.~A., Pereira, A.~A., Binney, J., Somers, T., and Sukhatme, G.~S.
  (2016).
\newblock Learning uncertainty in ocean current predictions for safe and
  reliable navigation of underwater vehicles.
\newblock {\em Journal of Field Robotics}, 33(1):47--66.

\bibitem[Hollinger et~al., 2013]{hollinger2013learning}
Hollinger, G.~A., Pereira, A.~A., and Sukhatme, G.~S. (2013).
\newblock Learning uncertainty models for reliable operation of autonomous
  underwater vehicles.
\newblock In {\em Robotics and Automation (ICRA), 2013 IEEE International
  Conference on}, pages 5593--5599. IEEE.

\bibitem[Huck et~al., 2012]{HuHoChLy2012}
Huck, S., Hokayem, P., Chatterjee, D., and Lygeros, J. (2012).
\newblock Stochastic localization of sources using autonomous underwater
  vehicles.
\newblock In {\em American Control Conference (ACC), 2012}, pages 4192--4197.

\bibitem[Laporte, 1992]{LAPORTE1992231}
Laporte, G. (1992).
\newblock The traveling salesman problem: An overview of exact and approximate
  algorithms.
\newblock {\em European Journal of Operational Research}, 59(2):231 -- 247.

\bibitem[Leonard et~al., 2010]{LPDDLZ_JFR10}
Leonard, N.~E., Paley, D.~A., Davis, R.~E., Fratantoni, D.~M., Lekien, F., and
  Zhang, F. (2010).
\newblock Coordinated control of an underwater glider fleet in an adaptive
  ocean sampling field experiment in monterey bay.
\newblock {\em Journal of Field Robotics}, 27(6):718--740.

\bibitem[Li and Guo, ]{LiGu2012}
Li, S. and Guo, Y.
\newblock Distributed source seeking by cooperative robots: All-to-all and
  limited communications.
\newblock In {\em Robotics and Automation (ICRA), 2012 IEEE International
  Conference on}, pages 1107--1112.

\bibitem[Lichtenstern, 2013]{lichtenstern2013kriging}
Lichtenstern, A. (2013).
\newblock Kriging methods in spatial statistics.
\newblock {\em Technische Universit{\"a}t M{\"u}nchen}.

\bibitem[Liu et~al., 2016]{gps2016urban}
Liu, X., Xi, T., and Ngai, E. (2016).
\newblock Data modelling with gaussian process in sensor networks for urban
  environmental monitoring.
\newblock In {\em Proc. 24th International Symposium on Modeling, Analysis and
  Simulation of Computer and Telecommunication Systems}, pages 457--462. IEEE
  Computer Society.

\bibitem[Lloret et~al., 2009]{lloret2009wireless}
Lloret, J., Garcia, M., Bri, D., and Sendra, S. (2009).
\newblock A wireless sensor network deployment for rural and forest fire
  detection and verification.
\newblock {\em sensors}, 9(11):8722--8747.

\bibitem[Low, 2009]{Low2009thesis}
Low, K.~H. (2009).
\newblock {\em Multi-robot Adaptive Exploration and Mapping for Environmental
  Sensing Applications}.
\newblock PhD thesis, Carnegie Mellon University, Pittsburgh, PA, USA.

\bibitem[Low et~al., 2011]{Low2011}
Low, K.~H., Dolan, J.~M., and Khosla, P. (2011).
\newblock Active markov information-theoretic path planning for robotic
  environmental sensing.
\newblock In {\em Proceedings of the 10th International Conference on
  Autonomous Agents and MultiAgent Systems (AAMAS-11)}, pages 753--760.

\bibitem[Ma et~al., 2016a]{ma2016information}
Ma, K.-C., Liu, L., and Sukhatme, G.~S. (2016a).
\newblock An information-driven and disturbance-aware planning method for
  long-term ocean monitoring.
\newblock In {\em IEEE/RSJ International Conference on Intelligent Robots and
  Systems}.

\bibitem[Ma et~al., 2016b]{ma2016DARS}
Ma, K.-C., Liu, L., and Sukhatme, G.~S. (2016b).
\newblock Multi-robot informative and adaptive planning for persistent
  environmental monitoring.
\newblock In {\em International Symposium on Distributed Autonomous Robotic
  Systems (DARS)}.

\bibitem[Ma et~al., 2017]{ma2017icra}
Ma, K.-C., Liu, L., and Sukhatme, G.~S. (2017).
\newblock Informative planning and online learning with sparse gaussian
  processes.
\newblock In {\em IEEE International Conference on Robotics and Automation
  (ICRA)}.

\bibitem[Maciel et~al., 2010]{MaBoPe2010}
Maciel, B., Borges~de Sousa, J., and Pereira, F. (2010).
\newblock Chemical plume source localization with multiple autonomous
  underwater vehicles.
\newblock In {\em OCEANS 2010 IEEE - Sydney}, pages 1--10.

\bibitem[Mayhew et~al., 2007]{MaSaTe2007}
Mayhew, C., Sanfelice, R., and Teel, A. (2007).
\newblock Robust source-seeking hybrid controllers for autonomous vehicles.
\newblock In {\em American Control Conference, 2007. ACC '07}, pages
  1185--1190.

\bibitem[Meliou et~al., 2007]{Meliou07}
Meliou, A., Krause, A., Guestrin, C., and Hellerstein, J.~M. (2007).
\newblock Nonmyopic informative path planning in spatio-temporal models.
\newblock In {\em Proceedings of National Conference on Artificial Intelligence
  (AAAI)}, pages 602--607.

\bibitem[Miles et~al., 2015]{Miles2015}
Miles, T., Lee, S.~H., Wåhlin, A., Ha, H.~K., Kim, T.~W., Assmann, K.~M., and
  Schofield, O. (2015).
\newblock Glider observations of the dotson ice shelf outflow.
\newblock {\em Deep Sea Research Part II: Topical Studies in Oceanography}.

\bibitem[Mukhopadhyay et~al., 2014]{mukhopadhyay2014collaborative}
Mukhopadhyay, S., Wang, C., Patterson, M., Malisoff, M., and Zhang, F. (2014).
\newblock Collaborative autonomous surveys in marine environments affected by
  oil spills.
\newblock In {\em Cooperative Robots and Sensor Networks 2014}, pages 87--113.
  Springer.

\bibitem[Neal, 1996]{neal1996bayesian}
Neal, R.~M. (1996).
\newblock Bayesian learning for neural networks.

\bibitem[Nguyen-tuong and Peters, 2008]{Nguyen-tuong08localgaussian}
Nguyen-tuong, D. and Peters, J. (2008).
\newblock Local gaussian process regression for real time online model learning
  and control.
\newblock In {\em In Advances in Neural Information Processing Systems 22
  (NIPS}.

\bibitem[Nuske et~al., 2011]{nuske2011yield}
Nuske, S., Achar, S., Bates, T., Narasimhan, S., and Singh, S. (2011).
\newblock Yield estimation in vineyards by visual grape detection.
\newblock In {\em 2011 IEEE/RSJ International Conference on Intelligent Robots
  and Systems}, pages 2352--2358. IEEE.

\bibitem[Ogren et~al., 2004]{OrFiLe2004}
Ogren, P., Fiorelli, E., and Leonard, N. (2004).
\newblock Cooperative control of mobile sensor networks:adaptive gradient
  climbing in a distributed environment.
\newblock {\em Automatic Control, IEEE Transactions on}, 49(8):1292--1302.

\bibitem[Oliveira and Rodrigues, 2011]{Oliveira11}
Oliveira, L.~M. and Rodrigues, J.~J. (2011).
\newblock Wireless sensor networks: a survey on environmental monitoring.
\newblock {\em Journal of Communications}, 6:143--151.

\bibitem[Opper, 1998]{Opper:1999:BAO:304710.304756}
Opper, M. (1998).
\newblock On-line learning in neural networks.
\newblock chapter A Bayesian Approach to On-line Learning, pages 363--378.
  Cambridge University Press, New York, NY, USA.

\bibitem[Ouyang et~al., 2014]{Ouyang2014MAS}
Ouyang, R., Low, K.~H., Chen, J., and Jaillet, P. (2014).
\newblock Multi-robot active sensing of non-stationary gaussian process-based
  environmental phenomena.
\newblock In {\em Proceedings of the 2014 International Conference on
  Autonomous Agents and Multi-agent Systems}, pages 573--580.

\bibitem[Paley et~al., 2008]{PaleyLeoZhang2006}
Paley, D.~A., Zhang, F., Fratantoni, D.~M., and Leonard, N.~E. (2008).
\newblock Glider control for ocean sampling: The glider coordinated control
  system.
\newblock {\em IEEE Transaction on Control System Technology}, 16(4):735--744.

\bibitem[Press et~al., 1996]{press1996numerical}
Press, W.~H., Teukolsky, S.~A., Vetterling, W.~T., and Flannery, B.~P. (1996).
\newblock {\em Numerical recipes in C}, volume~2.
\newblock Cambridge university press Cambridge.

\bibitem[Ranganathan et~al., 2011]{Ranganathan11}
Ranganathan, A., Yang, M.-H., and Ho, J. (2011).
\newblock {Online Sparse Gaussian Process Regression and Its Applications}.
\newblock {\em IEEE Transactions on Image Processing}, 20(2):391--404.

\bibitem[Rasmussen and Williams, 2005]{Rasmussen2005}
Rasmussen, C.~E. and Williams, C. K.~I. (2005).
\newblock {\em Gaussian Processes for Machine Learning}.
\newblock The MIT Press.

\bibitem[Shchepetkin and {McWilliams}, 2005]{shchepetkin_regional_2005}
Shchepetkin, A.~F. and {McWilliams}, J.~C. (2005).
\newblock {The regional oceanic modeling system (ROMS): a split-explicit,
  free-surface, topography-following-coordinate oceanic model}.
\newblock {\em Ocean Modelling}, 9(4):347--404.

\bibitem[Singh et~al., 2007]{Singh2007}
Singh, A., Krause, A., Guestrin, C., Kaiser, W., and Batalin, M. (2007).
\newblock Efficient planning of informative paths for multiple robots.
\newblock In {\em Proceedings of the 20th International Joint Conference on
  Artifical Intelligence}, IJCAI'07, pages 2204--2211.

\bibitem[Smith et~al., 2011]{smith2011persistent}
Smith, R.~N., Schwager, M., Smith, S.~L., Jones, B.~H., Rus, D., and Sukhatme,
  G.~S. (2011).
\newblock Persistent ocean monitoring with underwater gliders: Adapting
  sampling resolution.
\newblock {\em Journal of Field Robotics}, 28(5):714--741.

\bibitem[Smola and Bartlett, 2001]{smola2001sparse}
Smola, A.~J. and Bartlett, P.~L. (2001).
\newblock Sparse greedy gaussian process regression.
\newblock In {\em Advances in neural information processing systems}, pages
  619--625.

\bibitem[Soltero et~al., 2012]{SolteroSR12}
Soltero, D.~E., Schwager, M., and Rus, D. (2012).
\newblock Generating informative paths for persistent sensing in unknown
  environments.
\newblock In {\em IROS}, pages 2172--2179.

\bibitem[Stachniss et~al., 2008]{stachniss2008gas}
Stachniss, C., Plagemann, C., Lilienthal, A.~J., and Burgard, W. (2008).
\newblock Gas distribution modeling using sparse gaussian process mixture
  models.
\newblock In {\em Robotics: Science and Systems}, volume~3.

\bibitem[Tokekar et~al., 2010]{tokekar2010robotic}
Tokekar, P., Bhadauria, D., Studenski, A., and Isler, V. (2010).
\newblock A robotic system for monitoring carp in minnesota lakes.
\newblock {\em Journal of Field Robotics}, 27(6):779--789.

\bibitem[Trincavelli et~al., 2008]{Trincavelli08}
Trincavelli, M., Reggente, M., Coradeschi, S., Loutfi, A., Ishida, H., and
  Lilienthal, A.~J. (2008).
\newblock Towards environmental monitoring with mobile robots.
\newblock In {\em IEEE/RSJ International Conference on Intelligent Robots and
  Systems}, pages 2210--2215.

\bibitem[Watts et~al., 2012]{classification12}
Watts, A.~C., Ambrosia, V.~G., and Hinkley, E.~A. (2012).
\newblock {Unmanned Aircraft Systems in Remote Sensing and Scientific Research:
  Classification and Considerations of Use}.
\newblock {\em Remote Sensing}, 4(6):1671--1692.

\bibitem[Werner-Allen et~al., 2006]{werner2006fidelity}
Werner-Allen, G., Lorincz, K., Johnson, J., Lees, J., and Welsh, M. (2006).
\newblock Fidelity and yield in a volcano monitoring sensor network.
\newblock In {\em Proceedings of the 7th symposium on Operating systems design
  and implementation}, pages 381--396. USENIX Association.

\bibitem[Wu et~al., 2013]{WuChZh2013}
Wu, W., Chang, D., and Zhang, F. (2013).
\newblock A bio-inspired robust 3d plume tracking strategy using mobile sensor
  networks.
\newblock In {\em Decision and Control (CDC), 2013 IEEE 52nd Annual Conference
  on}, pages 4571--4578.

\bibitem[Yang, 2012]{yang2012high}
Yang, C. (2012).
\newblock A high-resolution airborne four-camera imaging system for
  agricultural remote sensing.
\newblock {\em Computers and electronics in agriculture}, 88:13--24.

\bibitem[Yu et~al., 2014]{YuSchRus14ICRA}
Yu, J., Schwager, M., and Rus, D. (2014).
\newblock Correlated orienteering problem and its application to informative
  path planning for persistent monitoring tasks.
\newblock In {\em IEEE/RSJ International Conference on Intelligent Robots and
  Systems}.

\end{thebibliography}
}

\end{document}